\documentclass[10pt,twocolumn,letterpaper]{article}

\usepackage{cvpr}
\usepackage{times}
\usepackage{epsfig}
\usepackage{graphicx}
\usepackage{amsmath}
\usepackage{amssymb}
\usepackage{tabularx}
\usepackage{balance}
\usepackage{multibib}

\usepackage[super]{nth}
\usepackage{booktabs}       \usepackage{graphicx}
\usepackage{amsmath,amssymb} \usepackage{color}
\usepackage[format=plain,labelformat=simple,labelsep=period,font=small,skip=6pt,compatibility=false]{caption}
\usepackage[font=footnotesize,skip=3pt]{subcaption}
\usepackage{wrapfig}
\usepackage{multirow}
\usepackage{array}

\definecolor{mygreen}{rgb}{0.2, 0.7, 0.1}

\usepackage[pagebackref=true,breaklinks=true,colorlinks,citecolor={mygreen},bookmarks=false]{hyperref}

\DeclareMathOperator*{\argmax}{arg\,max}
\DeclareMathOperator{\Tr}{tr}

\usepackage[ruled,vlined]{algorithm2e}
\usepackage[capitalize]{cleveref}
\crefname{section}{Sec.}{Sec.}

\usepackage{xspace}
\newcommand{\iid}{\emph{i.\thinspace{}i.\thinspace{}d.}\@\xspace}

\newcommand{\myparagraph}[1]{\medskip\noindent\textbf{#1}\hspace{0.25em}}

\newcommand{\thesupptitle}{Actor-Critic Instance Segmentation\\\large -- Supplemental Material --}
\newcommand{\thesuppauthor}{Nikita Araslanov$^\ast$ \qquad\quad Constantin A.~Rothkopf$^{\,\dagger,\S}$ \qquad\quad Stefan Roth$^{\ast,\S}$\\
$^\ast$Dept.~of Computer Science \quad $^\dagger$Institute of Psychology \quad $^\S$Centre for Cognitive Science\\
TU Darmstadt
}

\cvprfinalcopy 

\def\cvprPaperID{3709} 

\ifcvprfinal\fi

\newcites{supp}{References}

\usepackage{fancyhdr}
\usepackage{setspace}

\begin{document}

\title{Actor-Critic Instance Segmentation}

\author{Nikita Araslanov$^\ast$ \qquad\quad Constantin A.~Rothkopf$^{\,\dagger,\S}$ \qquad\quad Stefan Roth$^{\ast,\S}$\\
$^\ast$Dept.~of Computer Science \quad $^\dagger$Institute of Psychology \quad $^\S$Centre for Cognitive Science\\
TU Darmstadt
}

\maketitle

\pagestyle{empty}
\thispagestyle{fancy}

\begin{abstract}

Most approaches to visual scene analysis have emphasised parallel processing of the image elements.
However, one area in which the sequential nature of vision is apparent, is that of segmenting multiple, potentially similar and partially occluded objects in a scene. In this work, we revisit the recurrent formulation of this challenging problem in the context of reinforcement learning.
Motivated by the limitations of the global max-matching assignment of the ground-truth segments to the recurrent states,
we develop an actor-critic approach in which
the \emph{actor} recurrently predicts one instance mask at a time and utilises the gradient from a concurrently trained \emph{critic} network.
We formulate the state, action, and the reward such as to let the critic model long-term effects of the current prediction and incorporate this information into the gradient signal. 
Furthermore, to enable effective exploration in the inherently high-dimensional action space of instance masks, we learn a compact representation using a conditional variational auto-encoder. 
We show that our actor-critic model consistently provides accuracy benefits over the recurrent baseline on standard instance segmentation benchmarks. \end{abstract}

\section{Introduction}

Methods for instance segmentation have for the most part relied on the idea of parallel processing of the image elements and features within images~\cite{he2017mask}.
However, previous work~\cite{ren2016end,romera2016recurrent} suggests that instance segmentation can be formulated as a sequential visual task, akin
to human vision, for which substantial evidence has revealed that many vision tasks beyond eye movements are solved sequentially~\cite{ullman1987visual}.
While the segmentation accuracy of feed-forward pipelines hinges on a large number of object proposals, 
proposal-free recurrent models have a particular appeal for instance segmentation where the number of instances is unknown.
Also, the temporal context can facilitate a certain order of prediction: segmenting ``hard'' instances can be improved by conditioning on the masks of ``easy'' instances segmented first (\eg, due to occlusions, ambiguities in spatial context etc.; \cite{li2017not}).

A pivotal question of a recurrent formulation for instance segmentation is the assignment of the ground-truth segments to timesteps, since the order in which they have to be predicted is unknown.
Previously this was addressed using the Kuhn-Munkres algorithm~\cite{kuhn1955hungarian}, computing the max-matching assignment.
We provide some insight, however, that the final prediction ordering depends on the initial assignment.
Furthermore, the loss for every timestep is not informative in terms of its effect on future predictions.
Intuitively, considering the future loss for the predictions early on should improve the segmentation accuracy at the later timesteps.
Although this can be achieved by unrolling the recurrent states for gradient backpropagation,
such an approach quickly becomes infeasible for segmentation networks due to high memory demands.

In the past years, reinforcement learning (RL) has been showing promise in solving increasingly complex tasks~\cite{lillicrap2015continuous,mnih2013playing,mnih2016asynchronous}.
However, relatively little work has explored applications of RL outside its conventional domain, which we attribute to two main factors:
\emph{(1)} computer vision problems often lack the notion of the environment, which provides the interacting agent with the reward feedback;
\emph{(2)} actions in the space of images are often prohibitively high-dimensional, leading to tough computational challenges.

Here, we use an actor-critic (AC) model~\cite{barto1983neuronlike} to make progress regarding both technical issues of a recurrent approach to instance segmentation.
We use exploration noise to reduce the influence of the initial assignments on the segmentation ordering.
Furthermore, we design a reward function that accounts for the future reward in the objective function for every timestep.
Our model does not use bounding boxes -- often criticised due to their coarse representation of the objects' shape.
Instead, we built on an encoder-decoder baseline that makes pixelwise predictions directly at the scale of the input image.
To enable the use of RL for instance segmentation with its associated high-dimensional output space, we propose to learn a compact action-space representation through the latent variables of a conditional variational auto-encoder~\cite{kingma2013auto}, which we integrate into a recurrent prediction pipeline.

Our experiments demonstrate that our actor-critic model improves the prediction quality over its baseline trained with the max-matching assignment loss, especially at the later timesteps, and performs well on standard instance segmentation benchmarks. 
\pagestyle{plain}
\section{Related Work}

Instance segmentation has received growing attention in the recent literature.
One family of approaches focuses on learning explicit instance encodings~\cite{bai2016deep,de2017semantic,kirillov2016instancecut,liu2017sgn,uhrig2016pixel}, which are then clustered into individual instance masks using post-processing.
Another common end-to-end approach is to first predict a bounding box for each instance using dynamic pooling and then to produce a mask of the dominant object within the box using a separate segmentation network~\cite{he2017mask,li2016fully}.
These methods are currently best-practice, which can be attributed to the maturity of deep network-based object detection pipelines.
However, this strategy is ultimately limited by the detection performance, proposal set, and the need of additional processing to account for pixel-level context~\cite{arnab2016bottom,dai2016binstance}.

Making the predictions sequentially points at an alternative line of work.
Romera-Paredes \& Torr~\cite{romera2016recurrent} used a convolutional LSTM~\cite{xingjian2015convolutional} with a spatial softmax, which works well for isotropic object shapes and moderate scale variation.
At each timestep, the recurrent model of Ren \& Zemel~\cite{ren2016end} predicts a box location and scale for one instance.
However, the extent of the available context for subsequent segmentations is limited by the box.
Some benefits of the temporal and spatial context have been also re-asserted on the task of object detection~\cite{cheng17spatial,li2017sequential} and, much earlier, on image generation~\cite{gregor15dgrw} and recognition~\cite{larochelle2010learning}.
In contrast to these works, our method obviates the need for the intermediate bounding box representation and predicts masks directly at the image resolution.

We cast the problem as a sequential decision process, as is studied by reinforcement learning (RL; \cite{sutton1998reinforcement}).
Using the actor-critic framework~\cite{barto1983neuronlike}, we define the actor as the model that sequentially produces instance masks, whereas the critic learns to provide a score characterising the actor's performance.
Leveraging this score, the actor can be trained to improve the quality of its predictions.
This is reminiscent of the more recent Generative Adversarial Networks (GANs; \cite{goodfellow2014generative}), in which a generator relies on a discriminator to improve.
In particular, our model is similar to Wasserstein GANs~\cite{arjovsky2017wasserstein} in that the discriminator is trained on a regression-like loss, and to SeqGAN~\cite{yu2017seqgan} in that the generator's predictions are sequential.

One obstacle is the action dimensionality, since the sampling complexity required for exploration grows exponentially with the size of the actions.
A naive action representation for dense pixelwise predictions would lead to an action space of dimension in the order of $O(2^{H \times W})$ for images with resolution $H \times W$.
This is significantly higher than the action spaces of standard problems studied by reinforcement learning (usually, between 1 and 20), or even its applications to natural language processing~\cite{Bahdanau:2017:AC,Ranzato:2016:SLT}.
To address this, we suggest learning a compact representation using variational auto-encoders~\cite{kingma2013auto} to enable the crucial reduction of the problem from a high-dimensional discrete to a lower-dimensional continuous action space.
 
\section{Motivation}

As discussed above, we follow previous work in modelling instance segmentation as a sequential decision problem \cite{ren2016end,romera2016recurrent}, yielding one instance per timestep $t$.

\begin{figure}[t]
\centering
\includegraphics[width=0.47\textwidth]{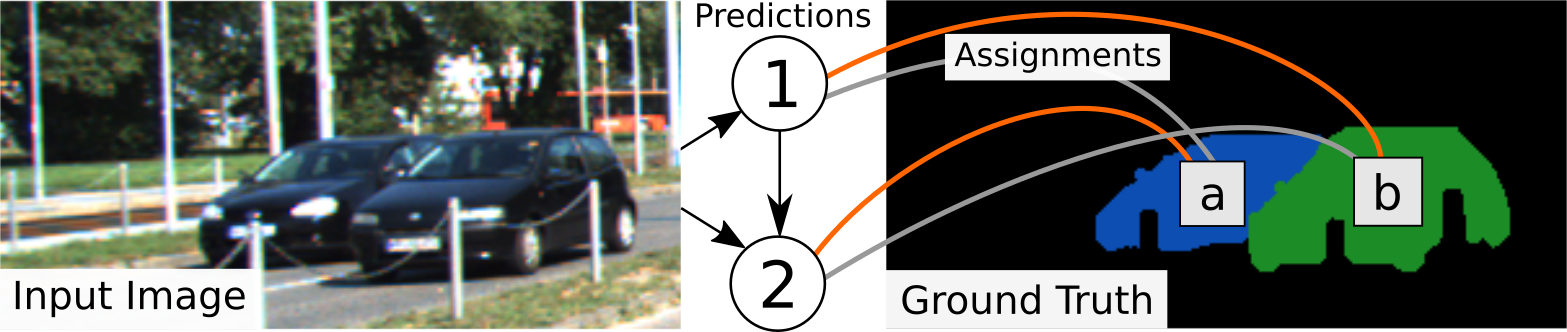}
\caption{\emph{Illustration of the max-matching assignment for instance segmentation.}
Consider an image with ground-truth instances \textit{a} and \textit{b}, and a recurrent model making predictions \textit{1} and \textit{2}.
In the constructed bipartite graph, each edge is assigned a weight corresponding to the IoU of the prediction with the connected ground truth.
From the set of possible assignments, depicted by the orange and grey edges,  max-matching finds the one that maximizes the sum of the IoUs.
The loss is then computed independently for each timestep \wrt to this assignment to the ground truth. 
}
\label{figure:assignment}
\end{figure}

To motivate our work, we revisit the standard practice of using the Kuhn-Munkres algorithm~\cite{kuhn1955hungarian} to assign the ground-truth instances to the predictions of a recurrent model.
Let $\theta$ parametrise the model and $U_\theta \in \mathbb{R}^{n \times n}$ denote a matrix of elements $u_{ij}$ measuring the score of the $i^\text{th}$ prediction \wrt the $j^\text{th}$ ground truth (\eg, the IoU).
The Kuhn-Munkres algorithm finds a permutation matrix as a solution to the max-matching problem
\begin{equation}
\argmax_{P \in \mathcal{P}} \: \Tr(U_\theta P),
\label{eq:KuhnMunkres}
\end{equation}
where $\mathcal{P}$ is the set of $n$-dimensional permutation matrices, \ie such that for all $P \in \mathcal{P}$, we have $\sum_j P_{ij} = 1, \sum_i P_{ij} = 1$, $P_{ij} \in \{0, 1\}$.
Given a differentiable loss function $l_\theta(i,j)$ (\eg, the binary cross-entropy), the model parameters $\theta$ are then updated to minimise $\sum_{ij,P_{ij}=1} l_\theta(i, j)$.

Consider a simple case of two ground-truth segments, $a$ and $b$, illustrated in \cref{figure:assignment}.
Without loss of generality, assume that the initial (random) model parameters yield
$u_{1a} + u_{2b} < u_{1b} + u_{2a}$, \ie the sum of scores for segmenting instance $a$ first and $b$ second is lower than in the opposite order.
This implies that max-matching will perform a gradient update step maximising the second sum, \ie $u_{1b} + u_{2a}$, but not the first.
As a consequence, for the updated parameters, the score for the ordering $b \rightarrow a$ is likely to dominate also at the later iterations of training.\footnote{Note that a formal proof is likely non-trivial due to the stochastic nature of training.}

Previous work~\cite{li2017not,OrderMatters} suggests that sequential models are not invariant to the order of predictions, including object segments (\cf supplemental material).
The implication from the example above is that $\sup \; u_{1a} + u_{2b} \neq \sup \; u_{1b} + u_{2a}$ (the $\sup$ is \wrt $\theta$).
One conceivable remedy to alleviate the effect of the initial assignment is to introduce noise $\epsilon$ to the score matrix $U$ (\eg, \iid Gaussian), such that \cref{eq:KuhnMunkres} becomes
\begin{equation}
\argmax_{P \in \mathcal{P}} \: \Tr\big((U_\theta + \epsilon) P\big).
\end{equation}
However, the noise in the loss function will not account for the inherent causality of the temporal context in recurrent models: perturbation of one prediction affects the consecutive ones.

In this work, we consider a more principled approach to encourage exploration of different orderings.
We inject \emph{exploration noise} at the level of individual predictions made by an \emph{actor} network,
while a jointly trained \emph{critic} network keeps track of the long-term outcome of the early predictions.
This allows to include in the gradient to the actor not only the immediate loss,
 but also the contribution of the current prediction to the future loss.
We achieve this by re-formulating the instance segmentation problem in the RL framework, which we briefly introduce next. 
\section{Notation and Definitions}

In the following, we define the key concepts of Markov decision processes (MDPs) in the context of instance segmentation.
For a more general introduction, we refer to \cite{sutton1998reinforcement}.

We consider finite-horizon MDPs defined by the tuple $(S, A, T, r)$, where the state space $S$, the action space $A$, the state transition $T: S \times A \rightarrow S$, and the reward \mbox{$r: S \times A \rightarrow \mathbb{R}$} are defined as follows. 

\myparagraph{State.} The \emph{state} $s_t \in S$ of the recurrent system is a tuple of the input image (and its task-specific representations) and an aggregated mask, \ie $s_t = (I, M_t)$.
The mask $M_t$ simply accumulates previous instance predictions, which encourages the model to focus on yet unassigned pixels.
Including $I$ enables access to the original input at every timestep.

\myparagraph{Action.} To limit the dimensionality of the action space, we define the \emph{action} $a_t \in A$ in terms of a compact mask representation.
To achieve this, we pre-train a conditional variational auto-encoder (cVAE; \cite{kingma2013auto}) to reproduce segmentation masks.
As a result, the action $a_t \in A=\mathbb{R}^l$ is a continuous latent representation of a binary mask and has dimensionality $l \ll H\cdot W$, while the decoder $\mathcal{D}: \mathbb{R}^l \rightarrow \mathbb{R}^{H \times W}$ ``expands'' the latent code to a full-resolution mask.

\myparagraph{State transition.}
As implied by the state and action definitions above, the \emph{state transition}
\begin{equation}
T\big((I, M_t), a_t\big) = \big(I, \max(M_t, \mathcal{D}(a_t))\big) 
\label{eq:state_transition}
\end{equation}
uses a pixelwise max of the previous mask and the decoded action, \ie~integrating the currently predicted instance mask into the previously accumulated predictions.

\myparagraph{Reward.} We design the \emph{reward function} to measure the progress of the state transition towards optimising a certain segmentation criterion.
The building block of the reward is the \textit{state potential}~\cite{ng1999policy}, which we base on the max-matching assignment of the current predictions to the ground-truth segments, \ie
\begin{equation}
   \phi_t := \max_{k \in \mathcal{P}(N)} \sum_{i=1}^t \mathcal{F}(\mathcal{S}_i, \mathcal{T}_{k_i}),
   \label{eq:def_potential}
\end{equation}
where $\mathcal{T}_i$ and $\mathcal{S}_{1 \leq i \leq t}$ are the $N$ ground-truth masks and $t$ predicted masks; $\mathcal{P}(N)$ is a collection of all permutations of the set $\{1, 2, ..., N\}$.
$\mathcal{F}(\cdot, \cdot)$ denotes a distance between the prediction and a ground-truth mask and can be chosen with regard to the performance metric used by the specific benchmark (\eg, IoU, Dice, etc.).
We elaborate on these choices in the experimental section.

The state potential in \cref{eq:def_potential} allows us to define the reward as the difference between the potentials of subsequent states
\begin{equation}
r_t := \phi(s_{t+1}) - \phi(s_{t}).
\label{eq:def_reward}
\end{equation}
Note that since the $(t+1)^\text{st}$ prediction might re-order the optimal assignment (computed with the Kuhn-Munkres algorithm), our definition of the reward is less restrictive \wrt the prediction order compared to previous work~\cite{ren2016end,romera2016recurrent}, which enforces a certain assignment to compute the gradient.
Instead, our immediate reward allows to reason about the relative improvement of one set of predictions over another.
 
\section{Actor-Critic Approach}

\subsection{Overview}

\begin{figure*}[t]
\def\svgwidth{\textwidth}
\begingroup  \makeatletter  \providecommand\color[2][]{    \errmessage{(Inkscape) Color is used for the text in Inkscape, but the package 'color.sty' is not loaded}    \renewcommand\color[2][]{}  }  \providecommand\transparent[1]{    \errmessage{(Inkscape) Transparency is used (non-zero) for the text in Inkscape, but the package 'transparent.sty' is not loaded}    \renewcommand\transparent[1]{}  }  \providecommand\rotatebox[2]{#2}  \ifx\svgwidth\undefined    \setlength{\unitlength}{2519.225bp}    \ifx\svgscale\undefined      \relax    \else      \setlength{\unitlength}{\unitlength * \real{\svgscale}}    \fi  \else    \setlength{\unitlength}{\svgwidth}  \fi  \global\let\svgwidth\undefined  \global\let\svgscale\undefined  \makeatother  \begin{picture}(1,0.33)    \renewcommand{\familydefault}{\sfdefault}
    \sffamily
    \put(0,0){\includegraphics[width=\unitlength]{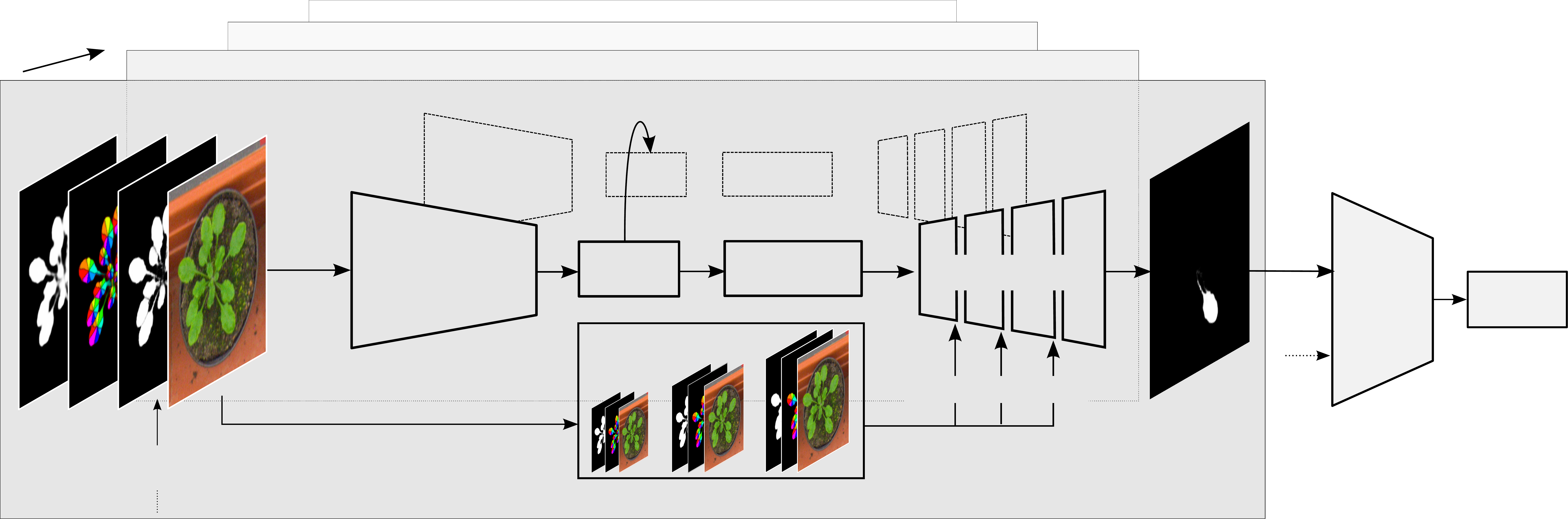}}    \put(0.365,0.25288867){\color[rgb]{0,0,0}\makebox(0,0)[lb]{\smash{$h_{t+1}$}}}    \put(0.085,0.25){\color[rgb]{0,0,0}\makebox(0,0)[lb]{\smash{State $s_t$}}}    \put(0.485,0.185){\color[rgb]{0,0,0}\makebox(0,0)[lb]{\smash{\scalebox{0.8}{Action}}}}    \put(0.468,0.15498913){\color[rgb]{0,0,0}\makebox(0,0)[lb]{\smash{\scalebox{0.75}{$a_t\!\sim\!\mathcal{N}(\mu, \sigma)$}}}}    \put(0.01569053,0.30577774){\color[rgb]{0,0,0}\makebox(0,0)[lb]{\smash{Time}}}    \put(0.944,0.13629859){\color[rgb]{0,0,0}\makebox(0,0)[lb]{\smash{\scalebox{0.65}{$Q(s_t,a_t)$}}}}    \put(0.3798,0.1537){\color[rgb]{0,0,0}\makebox(0,0)[lb]{\smash{\scalebox{0.8}{LSTM}}}}    \put(0.859,0.134165){\color[rgb]{0,0,0}\makebox(0,0)[lb]{\smash{Critic}}}    \put(0.248,0.151){\color[rgb]{0,0,0}\makebox(0,0)[lb]{\smash{Encoder}}}    \put(0.606,0.151){\color[rgb]{0,0,0}\makebox(0,0)[lb]{\smash{Decoder}}}    \put(0.601,0.078){\color[rgb]{0,0,0}\makebox(0,0)[lb]{\smash{\scriptsize Concatenate}}}    \put(0.378,0.107){\color[rgb]{0,0,0}\makebox(0,0)[lb]{\smash{\scalebox{0.75}{State Pyramid}}}}    \put(0.04311179,0.0280081){\color[rgb]{0,0,0}\makebox(0,0)[lb]{\smash{\scalebox{0.8}{$\max(M_{t-1}, m_{t-1})$}}}}    \put(0.82291545,0.11557518){\color[rgb]{0,0,0}\makebox(0,0)[lb]{\smash{$s_t$}}}    \put(0.72,0.05){\color[rgb]{0,0,0}\makebox(0,0)[lb]{\smash{Mask $m_t$}}}    \put(0.62,0.01){\color[rgb]{0,0,0}\makebox(0,0)[lb]{\smash{Actor}}}  \end{picture}\endgroup \caption{\emph{Actor-critic model for instance segmentation.} The model relies on two types of context: a spatial permutation-invariant state $s_t$ accumulates the masks, whereas the hidden LSTM state $h_t$ models a temporal context sensitive to the prediction ordering. The State Pyramid propagates the high-res information at multiple scales to the decoder to compensate the loss of resolution at the bottleneck section.}
\label{fig:model}
\end{figure*}
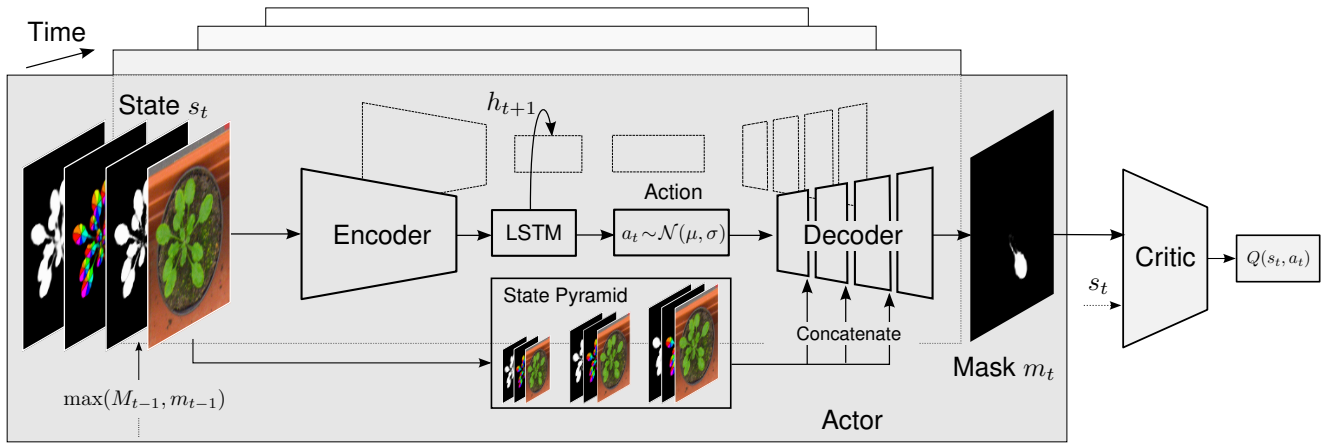

The core block of the actor model, shown in \cref{fig:model}, is a conditional variational auto-encoder (cVAE;~\cite{kingma2013auto}).
The encoder computes a compact vector of latent variables, encoding a full-resolution instance mask.
The decoder recovers such a mask from the latent code.
Using the transition function defined by~\cref{eq:state_transition}, the latest prediction updates the state, and the procedure repeats until termination.

The actor relies on two types of context with complementary properties.
As discussed above, the mask $M_t$ is a component of the state $s_t$, which accumulates the masks produced in the previous steps.
It provides permutation-invariant temporal context of high-resolution cues, encouraging the network to focus on yet unlabelled pixels.
The hidden state of actor $h_t$ is implemented by the LSTM~\cite{hochreiter1997long} at the bottleneck section and is unknown to the critic.
In contrast to state $s_t$, the representation of the hidden state is learned and can be sensitive to the prediction ordering due to the non-commutative updates of the LSTM state.
The hidden state, therefore, contributes to the temporal context and is shown to be particularly helpful for counting in the ablation study.

We train our model in two stages as described next.

\subsection{Pre-training}
\label{sec:model-pre-training}

We pre-train the actor cVAE to reconstruct the mask of the target segment.
The input to the network consists of the image and the binary mask of a randomly chosen ground-truth instance.
To account for the loss of high-resolution information at the latent level, the decoder is conditioned on the input image and auxiliary channels of instance-relevant representations supplied at multiple scales, which we term \emph{State Pyramid}.
One channel contains the foreground prediction, while the other 8 channels encode the instance angle quantisation of~\cite{uhrig2016pixel}, thereby binning the pixels of the object segment into quantised angles relative to the object's centroid.
These features assist in instance detection and disambiguation, since a neighbourhood of orthogonal quantisation vectors indicates occlusion boundaries and object centroids.
Following Ren \& Zemel~\cite{ren2016end}, we predict the angles with a pre-processing network~\cite{long2015fully} trained in a standalone fashion.

The auto-encoder uses the binary cross-entropy (BCE) as the reconstruction loss.
For the latent representation, we use a Gaussian prior with zero mean and unit variance.
The corresponding loss function is taken as the Kullback-Leibler divergence~\cite{kingma2013auto}.

\subsection{Training}
\label{sec:model-training}

During training, we learn a new encoder to sequentially predict segmentation masks.
In addition to the image, the encoder also receives the auxiliary channels used during pre-training.
In contrast, however, the encoder is encouraged to learn instance-sensitive features, since the decoder expects the latent code of ground-truth masks.

Algorithm~\ref{alg:opt} provides an outline of the training procedure.
The actor is trained jointly with the critic from a buffer of experience accumulated in the \emph{episode execution} step.
In the \emph{policy evaluation}, the critic is updated to minimise the error of approximating the expected reward, while in the \emph{policy iteration} the actor receives the gradient from the critic to maximise the Q-value.

\myparagraph{Episode execution.}
For an image with $N$ instances, we define the episode as the sequence of $N$ predictions.
The algorithm randomly selects a mini-batch of images without replacement and provides it as inputs to the actor.
Using the reparametrisation trick~\cite{kingma2013auto}, the actor samples an action corresponding to the next prediction of the instance mask.
The results of the predictions and corresponding rewards are saved in a buffer.
At the end of each episode, the target Q-value can be computed for each timestep $t$ as a sum of immediate rewards.

\myparagraph{Policy evaluation.}
The critic network parametrised by $\phi$, maintains an estimate of the Q-value defined as a function of the state and action guided by the policy $\mu$:
\begin{equation}
  Q_\phi(s_t, a_t) = \mathbb{E}_{a_j \sim \mu(s_j), j>t}  \Bigg[\sum_{i=t}^N \gamma^{i-t}  r_i (s_i, a_i) \Bigg].
\end{equation}
Note the finite sum in the expectation due to the finite number of instances in the image.
The critic's loss $\mathcal{L}_{\text{Critic},t}$ for timestep $t$ is defined as the squared $\mathcal{L}_2$-distance with a discounted sum of rewards
\begin{equation}
\begin{aligned}
  \mathcal{L}_{\text{Critic}, t} = \Bigg\| Q_\phi(s_t, a_t) - \sum_{i=t}^N \gamma^{i-t} r_i \Bigg\|_2^2,
\end{aligned}
\label{eq:critic_obj}
\end{equation}
where $\gamma \in (0, 1)$ is a \textit{discount factor} that controls the \emph{time horizon}, \ie the degree to which future rewards should be accounted for.
The hyperparameter $\gamma$ allows to trade off the time horizon for the difficulty of the reward approximation:
as $\gamma \rightarrow 1$, the time horizon extends to all states, but the critic has to approximate a distant future reward based only on the current state and action.
We update the parameters of the critic to minimise \cref{eq:critic_obj} using the samples of the state-actions and rewards in the buffer, and
set $\gamma = 0.9$ throughout our experiments.

\myparagraph{Policy iteration.}
The actor samples an action $a_t \in A$ from a distribution provided by the current policy $\mu_\theta: S \rightarrow A$, parametrised by $\theta$
and observes a reward $r_t$ computed by \cref{eq:def_reward}.
Given the initial state $s_1$, the actor's goal is to find the policy maximising the expected total reward,
$\theta^\ast = \argmax_\theta \: \mathbb{E}_{a_j \sim \mu_\theta(s_j)}  \big[\sum_{i=1}^N \gamma^i r_i (s_i, a_i) \big]$, approximated by the critic.
To achieve this, the state $s_t = (I, M_t)$ and the actor's mask prediction $m_t$ are passed to the critic, which produces the corresponding Q-value.
The gradient maximising the Q-value is computed via backpropagation and returned to the actor for its parameter update.

We found that fixing the decoder during training led to faster convergence.
Since the critic only approximates the true loss, its gradient is biased, which in practice can break the assumption we maintain during training -- that an optimal mask can be reconstructed from a normally-distributed latent space.
We fix the decoder and maintain the KL-divergence loss $\mathcal{L}_\text{KL}$ while sampling new actions, thus encouraging exploration of the action space.
In our ablation study, we verify that such exploration improves the segmentation quality.
Note that we do not pre-define the layout of the actions, but only maintain the Gaussian prior.

To further improve the stability of the joint actor-critic training, we use a \textit{warm-up phase} for the critic:
episode execution and update of the critic happen without updating the actor for a number of epochs.
This gives the critic the opportunity to adapt to the current action and state space of the actor.
We could confirm in our experiments that pre-training the decoder was crucial;
omitting this step resulted in near-zero rewards from which it proved to be difficult to train the critic even with the warm-up phase.

\begin{algorithm}
\footnotesize
  \DontPrintSemicolon
  \SetAlgoLined
  \SetKwData{Buffer}{buffer}
  \SetKwData{Episode}{episode}
  \SetKwData{Minibatch}{minibatch}
  \SetKwData{Epoch}{epoch}
  \SetKwData{NumEpochs}{NumEpochs}
  Initialise actor $\mu_\theta(s)$ from pre-training and critic $Q_\phi(s, a)$\;
  \For{$\Epoch=1,\NumEpochs$}
  {
    \ForEach{\Minibatch}
    {
      \tcp{\footnotesize accumulate \Buffer for replay}
      \Buffer $\leftarrow [\:]$\;
      \ForEach{(Image, $\{\mathcal{T}\}_{1,...,N}$) in \Minibatch}
      {
        Initialise mask $M_1 \leftarrow Empty$\;
        Initialise state $s_1 \leftarrow (Image, M_1)$\;
        \Episode $\leftarrow [\:]$\;
        \For{$t = 1, N$}
        {
          Sample action $a_t \sim \mu_\theta(s_t)$\;
          Obtain next state $s_{t+1}=T(s_t, a_t)$ with \cref{eq:state_transition}\;
          Add $(s_t, a_t, s_{t+1})$ to \Episode\;
        }
        Compute rewards for \Episode with \cref{eq:def_reward}\;
        Add \Episode with rewards to \Buffer\;
      }
      \tcp{\footnotesize Batch-update critic from \Buffer}
      \ForEach{$(s_t, a_t, r_t, s_{t+1})$ in \Buffer}
      {
        $\phi \leftarrow \phi - \alpha_{\text{critic}} \nabla_{\phi} \big(Q_\phi(s_t, a_t) - \sum_{i=t}^N \gamma^{i-t} r_i\big)^2$\;
      }
      \tcp{\footnotesize Batch-update actor using critic}
      Initialise states $s_1$ from \Buffer\; 
      \For{$t=1,N$}
      {
        Sample action $a_t \sim \mu_\theta(s_t)$\;
        $\theta\!\leftarrow\!\theta + \alpha_{\text{act}} \nabla_{a_t} Q_\phi(s_t, a_t) \nabla_{\theta} \mu_\theta(s_t) - \beta_{\text{act}} \nabla_{\theta} \mathcal{L}_\text{KL}$\!\;
        Obtain next state $s_{t+1}=T(s_t, a_t)$ using \cref{eq:state_transition}\;
      }
    }
  }
  \caption{Actor-critic training}
  \label{alg:opt}
\end{algorithm}

\myparagraph{Termination.}
We connect the hidden state $h_t$ and the last layer preceding it (via a skip connection) to a single unit predicting ``1'' to continue prediction, and ``0'' to indicate the terminal state.
Using the ground-truth number of instances, we train this unit with the BCE loss.

\myparagraph{Inference.}
We recurrently run the actor network until the termination prediction.
To obtain the masks, we discard the deviation part and only take the mean component of the action predicted by the encoder and pass it through the decoder.
We do not use the critic network at inference time.

\myparagraph{Implementation details.\footnote{Code is available at \url{https://github.com/visinf/acis/}.}}
We use a simple architecture similar to~\cite{romera2016recurrent} for both the critic and the actor networks trained with Adam~\cite{kingma14adam} until the training loss on validation data does not improve (\cf supplemental material).

\subsection{Discussion}

In the actor-critic model the critic plays the role of modelling the subsequent rewards for states $s_{i>t}$ given state $s_t$.
Hence, if the critic's approximation is exact, the backpropagation through time (BPTT; \cite{werbos1988generalization}) until the first state is not needed:
to train the actor, we need to compute the gradient \wrt the future rewards already predicted by the critic.
The implication of this property is that memory-demanding networks, such as those for dense prediction, can be effectively trained with truncated BPTT and the critic, even in case of long sequences.
Moreover, using the critic's approximation allows the reward be a non-differentiable, or even a discontinuous function tailored specifically to the task.
 
\section{Experiments}

In our experiments, we first quantitatively verify the importance of the different components in our model and investigate the sources of the accuracy benefits of the actor-critic over the baseline. 
Then, we use two standard datasets of natural images for the challenging task of instance segmentation, and compare to the state of the art.

\subsection{Ablation study}
\label{section:ablation}

We design a set of experiments to investigate the effect of various aspects of the model using 
the A1 benchmark of the \emph{Computer Vision Problems in Plant Phenotyping} (CVPPP) dataset~\cite{ScharrMFK0LLPPV16}.
It comprises a collection of 128 images of plants taken from a top view with leaf annotation as ground-truth instance masks.
We downsized the original 128 images in the training set by a factor of two and used a centre crop of size $224 \times 224$ for training.
For the purpose of the ablation study, we randomly select 103 images from the CVPPP A1 benchmark for training and report the results on the remaining 25 images.

To compute the reward for our actor-critic model (Eq.~\ref{eq:def_potential}), we use the Dice score computed as $\mathcal{F}(\mathcal{S}, \mathcal{T}) = \frac{2\sum_i{\mathcal{S}_i \mathcal{T}_i}}{\sum_i \mathcal{S}_i + \sum_i \mathcal{T}_i}$.
The dimensionality of the latent action space is fixed to 16.

In the first part of the experiment, we look into how different terms in the loss influence the segmentation quality, measured in Symmetric Best Dice (SBD), and the absolute Difference in Counting ($\mid$DiC$\mid$).
Specifically, we train five models:
\emph{BL} is an actor-only recurrent model trained with BPTT through all states.
We use the BCE loss and Dice-based max-matching as a heuristic for assigning the ground truth to predictions, similar to~\cite{ren2016end,romera2016recurrent}.
\emph{BL-Trunc} is similar to \emph{BL}, but is trained with a truncated, one-step BPTT.
We train our actor-critic model \emph{AC-Dice} with the gradient from the critic approximating the Dice score.
\emph{AC-Dice-NoKL} is similar to the \emph{AC-Dice} model, \ie the actor is trained jointly with the critic, but we remove the KL-divergence term, which encourages exploration, from the loss of the actor.
Lastly, we verify the benefit of the State Pyramid, the multi-res spatial information provided to the decoder, by comparing to a baseline without it (\emph{AC-Dice-NoSP}).

\begin{table}[!t]
\footnotesize
\centering
\begin{tabularx}{\columnwidth}{@{}Xcc@{}}
\toprule
Model & SBD $\uparrow$ & $\mid$DiC$\mid$ $\downarrow$ \\
\midrule
BL & 80.0 & 1.08\\ BL-Trunc & 79.4 & 1.32\\ AC-Dice & \textbf{80.5}  & \textbf{0.88} \\ AC-Dice-NoKL & 75.4 & 1.36\\ AC-Dice-NoSP & 61.3 & 1.52\\ \bottomrule
\end{tabularx}
\caption{\emph{Evaluation on CVPPP val.}
We compare our baseline with fully-unrolled (\emph{BL}) and truncated BPTT (\emph{BL-Trunc}) to the actor-critic with Dice-based reward, with (\emph{AC-Dice}) and without (\emph{AC-Dice-NoKL}) exploration, and without the State Pyramid (\emph{AC-Dice-NoSP}).}
\label{table:ablation}
\end{table}

\begin{table}[!t]
\footnotesize
\centering
\begin{tabularx}{\columnwidth}{@{}Xc@{\hskip 0.5em}cc@{\hskip 0.5em}cc@{\hskip 0.5em}c@{}}
\toprule
\multirow{2}{*}{Model} & \multicolumn{2}{c}{LSTM + Mask} & \multicolumn{2}{c}{Mask only} & \multicolumn{2}{c}{LSTM only} \\
\cmidrule(lr){2-3}
\cmidrule(lr){4-5}
\cmidrule(lr){6-7}
& Dice$^\ast$ $\uparrow$ & $\mid$DiC$\mid$ $\downarrow$ & Dice$^\ast$ $\uparrow$ & $\mid$DiC$\mid$ $\downarrow$ & Dice$^\ast$ $\uparrow$ & $\mid$DiC$\mid$ $\downarrow$ \\
\midrule
BL & 78.6 & 1.04 & 76.6 & 4.36 & 6.5 & 3.96\\
BL-Trunc & 77.9 & 1.72 & 77.5 & 6.24 & 6.0 & 4.8\\
AC-Dice & 78.4 & 0.88 & 78.5 & 1.92 & 5.8 & 4.36\\
\bottomrule
\multicolumn{7}{l}{$^\ast$ computed by max-matching and ground-truth stopping}\\
\end{tabularx}
\caption{\emph{Contribution of recurrent states to mask quality} measured by Dice and absolute Difference in Counting $\mid$DiC$\mid$ on CVPPP val.}
\label{table:ablation_state}
\end{table}

\begin{figure}[t]
\includegraphics[width=0.51\textwidth]{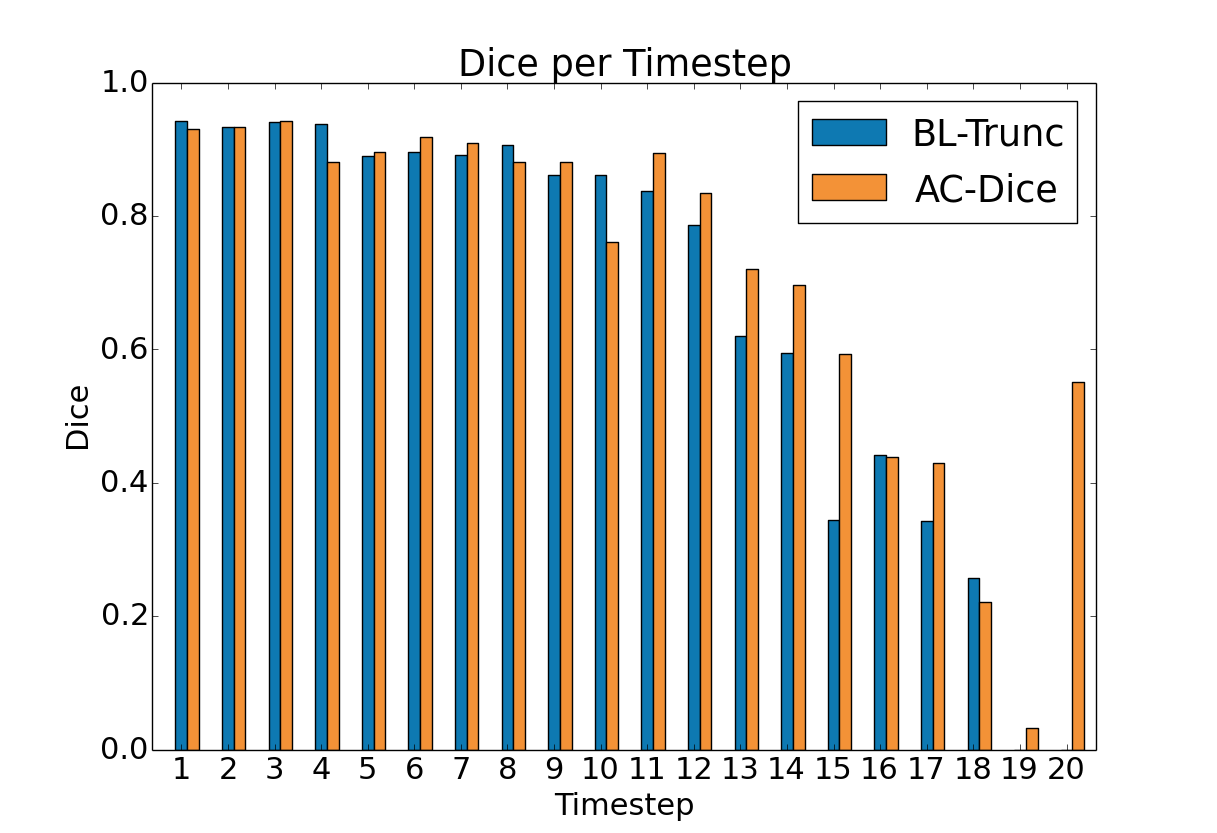}
\caption{\emph{Dice score} of our actor-critic model (\emph{AC-Dice}) \vs our baseline with truncated BPTT (\emph{BL-Trunc}) on CVPPP val, \emph{averaged for each timestep.}
We observe the advantage of our actor-critic model at later timesteps, which is an expected benefit of including the estimate of the expected reward in the loss at the earlier timesteps.
Note that few images contain 20 instances, hence a large variance for this timestep.}
\label{figure:dice_timesteps}
\end{figure}

The side-by-side comparison of these models summarised in Table~\ref{table:ablation} reveals that \emph{AC-Dice} exhibits a superior accuracy compared to the baselines, both in terms of Dice and counting.
Using the KL-divergence term in the loss improves the actor, which shows the value of action exploration in a consistent action space.
We also observed that training \emph{AC-Dice-NoKL} would sometimes diverge and require a restart with the critic warm-up.
Furthermore, the State Pyramid aids the decoder, as removing it leads to a significant drop in mask quality.
Surprisingly, \emph{BL-Trunc} is only slightly worse than \emph{BL}, which however has by far higher memory demands than both \emph{AC-Dice} and \emph{BL-Trunc} in the setting of long sequences and high resolutions.

To further investigate the accuracy gains of the actor-critic model, we report the average Dice score \wrt the corresponding timestep of the prediction in \cref{figure:dice_timesteps}.
The histogram confirms our intuition that incorporating the future reward into the loss function for every timestep, as modelled by the critic, should improve the segmentation quality at later stages of prediction:
the Dice score of the actor-critic model tends to be tangibly higher especially at the later timesteps.
Note that the contribution of this benefit to the average score across the dataset is moderated by not all images in the dataset having many instances.

In the next part of the experiment, we are interested in the reliance of the model on the recurrent state.
Recall that our model maintains the mask accumulating the previous predictions as well as the hidden LSTM state.
We alternately ``block'' either of the states by providing a zero tensor at every timestep.
We consider only the first $n$ predictions to compute the Dice score, where $n$ is the number of ground-truth masks.
We stop the iterations if no termination was predicted after 21 timesteps, since the largest number of instances in our validation set is 20.
The results in Table~\ref{table:ablation_state} show that the LSTM plays an important role for counting (or, termination prediction), while having almost no effect on the mask quality.
The networks have learned a sequential prediction strategy given only the binary mask of previously predicted pixels.
Note that in contrast to the baseline models, actor-critic training reduced the dependence on the LSTM state for counting (AC-Dice), 
which suggests that the actor makes a better use of the state mask to make the next prediction.

\subsection{Instance segmentation}

\begin{table}[t]
\centering
\footnotesize
\begin{tabularx}{\columnwidth}{@{}Xcc@{}}
\toprule
Model & SBD $\uparrow$ & $\mid$DiC$\mid$ $\downarrow$ \\
\midrule
RIS~\cite{romera2016recurrent} & 66.6 & 1.1\\ MSU~\cite{ScharrMFK0LLPPV16} & 66.7 & 2.3\\
Nottingham~\cite{ScharrMFK0LLPPV16} & 68.3 & 3.8\\
IPK~\cite{Pape:2014:3DH} & 74.4 & 2.6\\
DLoss~\cite{de2017semantic} & 84.2 & 1.0\\ E2E~\cite{ren2016end} & 84.9 & 0.8\\
\midrule
Ours (AC-Dice) & 79.1 & 1.12\\
\bottomrule
\end{tabularx}
\caption{\emph{Segmentation quality of our actor-critic model on CVPPP test} with Dice-based reward (\textit{AC-Dice}) in terms of Symmetric Best Dice (SBD) and absolute Difference in Counting ($\mid$DiC$\mid$).}
\label{table:results_cvppp}
\end{table}

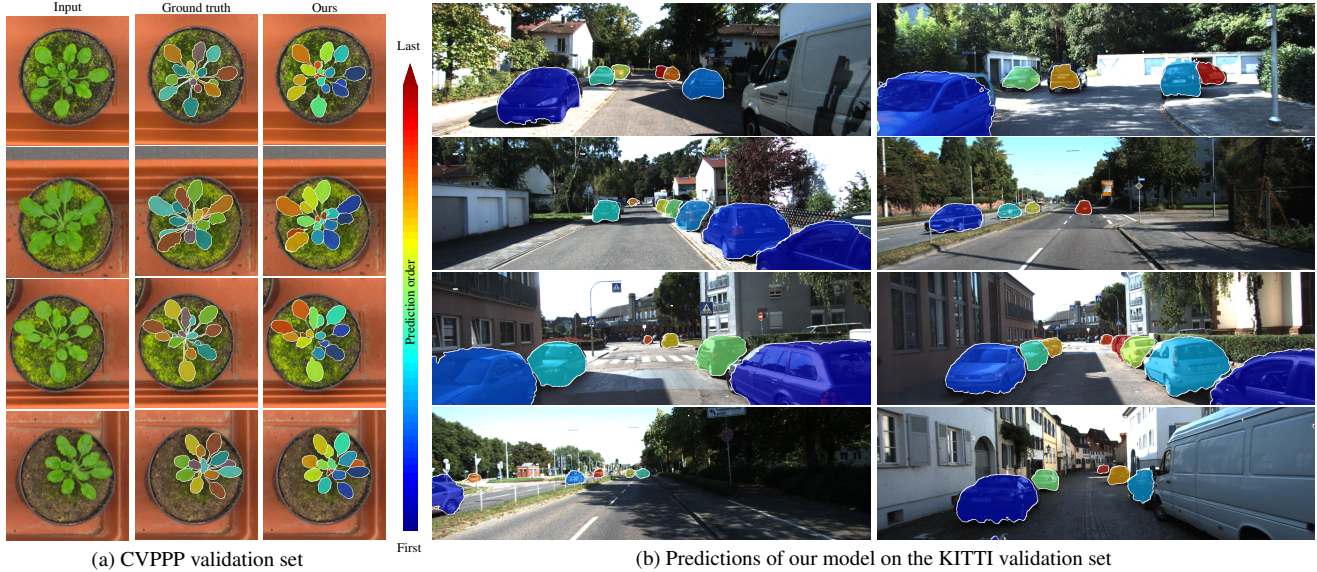
\begin{figure*}[t]
\begin{subfigure}{.29\textwidth}
  \begin{minipage}{.32\textwidth}
    \tiny
    \centering
    Input
    \includegraphics[width=\textwidth]{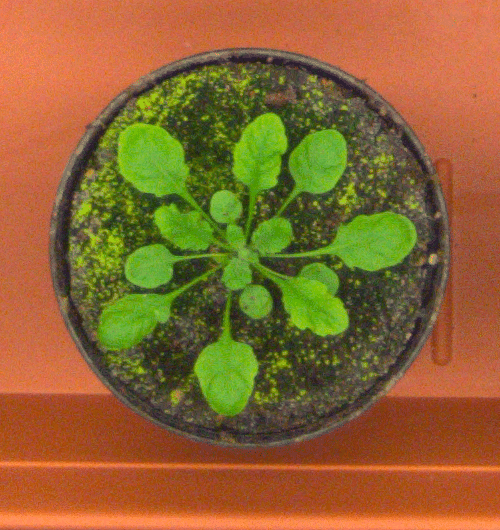}
    \includegraphics[width=\textwidth]{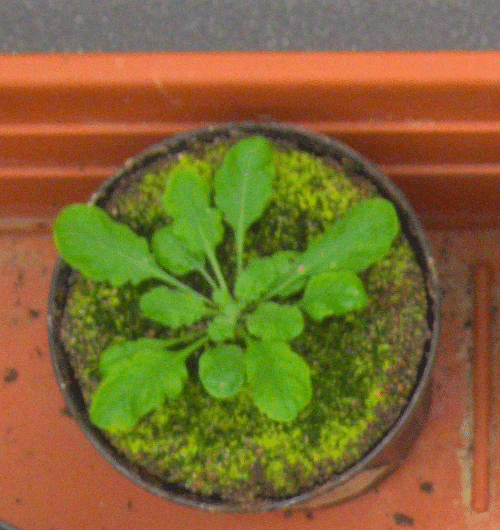}
    \includegraphics[width=\textwidth]{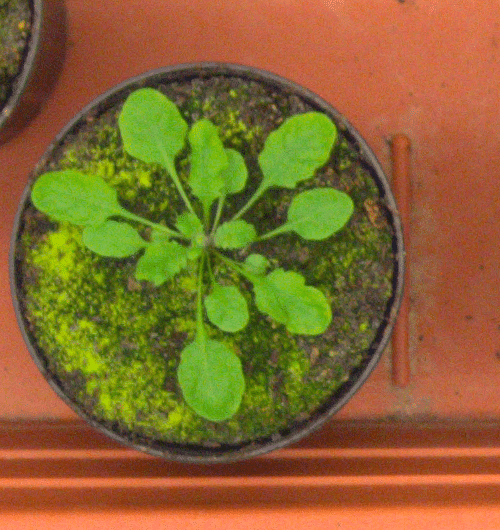}
    \includegraphics[width=\textwidth]{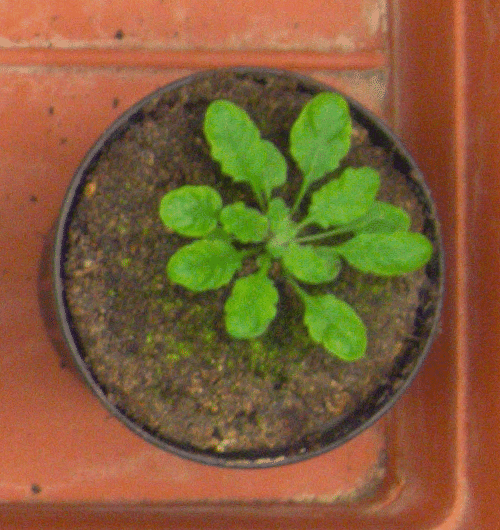}
  \end{minipage}
  \begin{minipage}{.32\textwidth}
    \tiny
    \centering
    Ground truth
    \includegraphics[width=\textwidth]{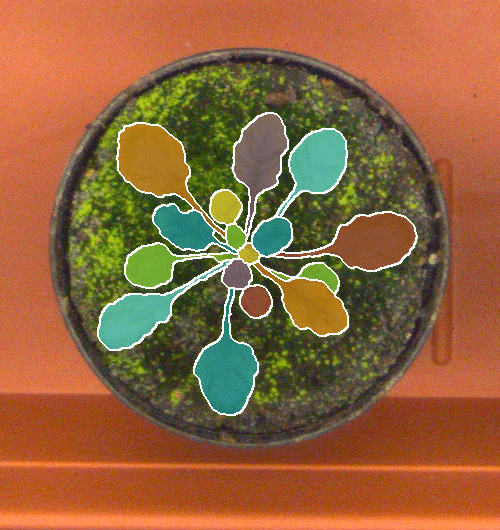}
    \includegraphics[width=\textwidth]{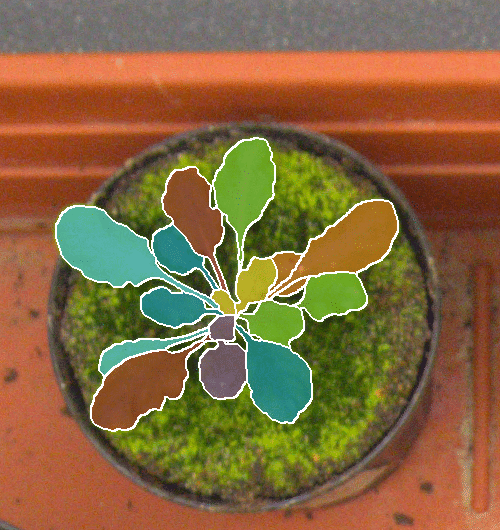}
    \includegraphics[width=\textwidth]{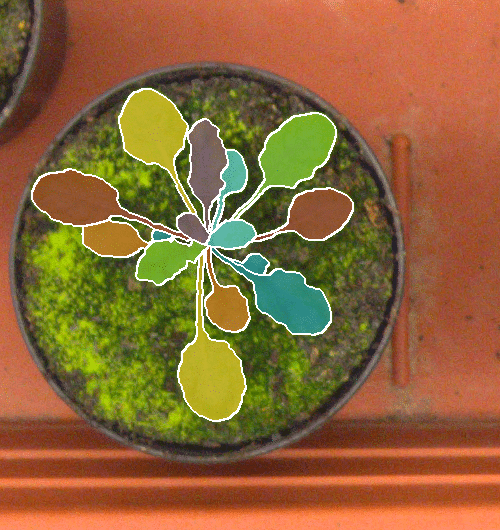}
    \includegraphics[width=\textwidth]{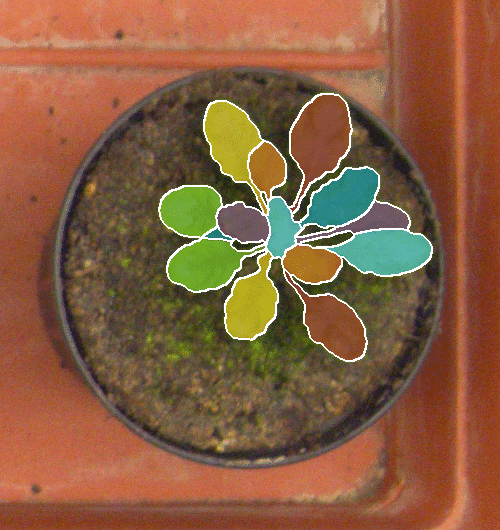}
  \end{minipage}
  \begin{minipage}{.32\textwidth}
    \tiny
    \centering
    Ours
    \includegraphics[width=\textwidth]{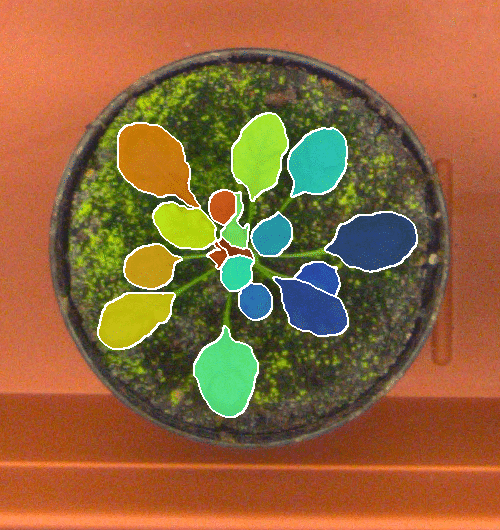}
    \includegraphics[width=\textwidth]{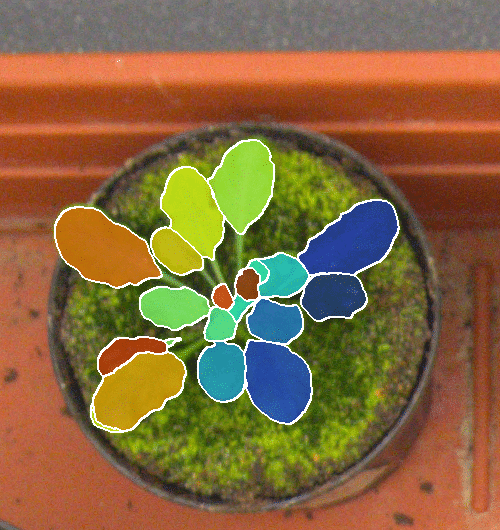}
    \includegraphics[width=\textwidth]{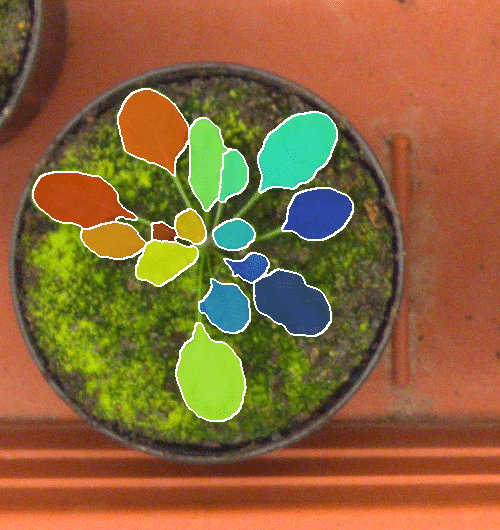}
    \includegraphics[width=\textwidth]{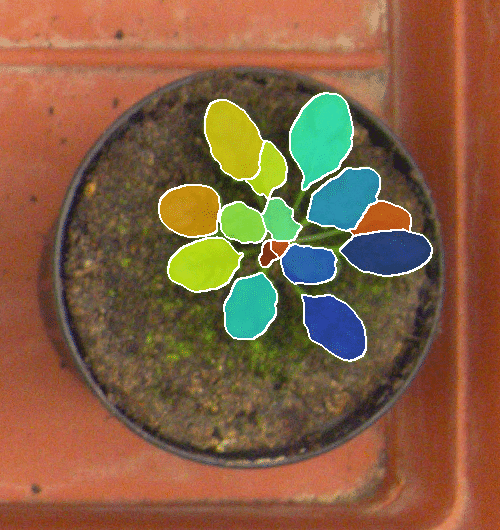}
  \end{minipage}
  \caption{CVPPP validation set}
  \label{figure:results_cvppp}
\end{subfigure}
\hspace{0.01em}
\begin{minipage}{0.018\textwidth}
\centering
\vspace{0.7em}
\def\svgwidth{\textwidth}
\begingroup  \makeatletter  \providecommand\color[2][]{    \errmessage{(Inkscape) Color is used for the text in Inkscape, but the package 'color.sty' is not loaded}    \renewcommand\color[2][]{}  }  \providecommand\transparent[1]{    \errmessage{(Inkscape) Transparency is used (non-zero) for the text in Inkscape, but the package 'transparent.sty' is not loaded}    \renewcommand\transparent[1]{}  }  \providecommand\rotatebox[2]{#2}  \ifx\svgwidth\undefined    \setlength{\unitlength}{27.78808594bp}    \ifx\svgscale\undefined      \relax    \else      \setlength{\unitlength}{\unitlength * \real{\svgscale}}    \fi  \else    \setlength{\unitlength}{\svgwidth}  \fi  \global\let\svgwidth\undefined  \global\let\svgscale\undefined  \makeatother  \begin{picture}(1,21.60873616)    \put(0,0){\includegraphics[width=\unitlength,page=1]{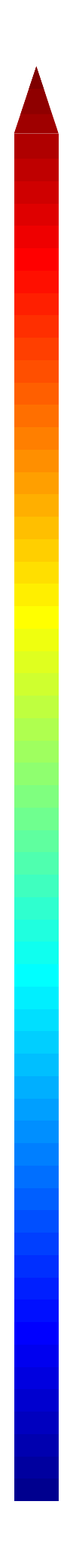}}    \put(0.68782376,8.77804252){\color[rgb]{0,0,0}\rotatebox{90}{\makebox(0,0)[lb]{\tiny Prediction order}}}    \put(-0.04944649,0.00713407){\color[rgb]{0,0,0}\makebox(0,0)[lb]{\tiny First}}    \put(-0.04354244,21.24145449){\color[rgb]{0,0,0}\makebox(0,0)[lb]{\tiny Last}}  \end{picture}\endgroup \end{minipage}\hspace{0.01em}
\begin{subfigure}{0.68\textwidth}
  \centering
  \begin{minipage}{.49\textwidth}
    \includegraphics[width=\textwidth]{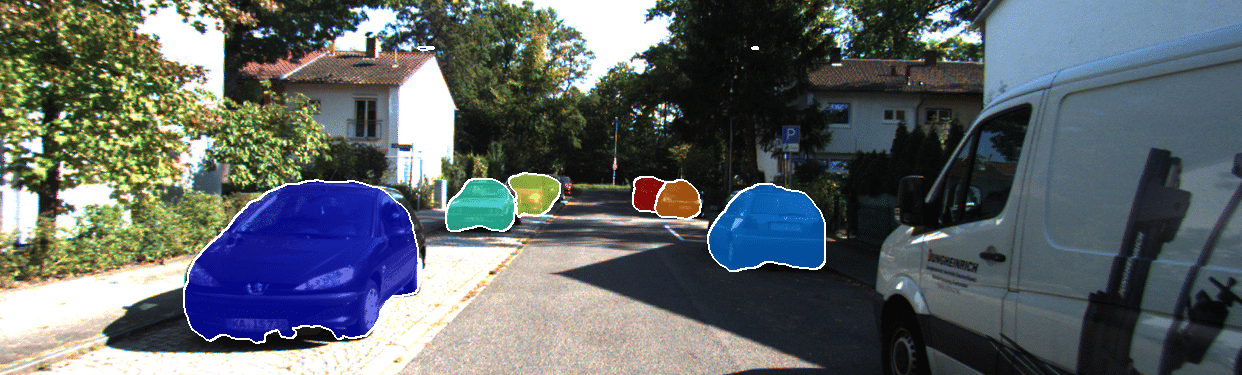}
    \includegraphics[width=\textwidth]{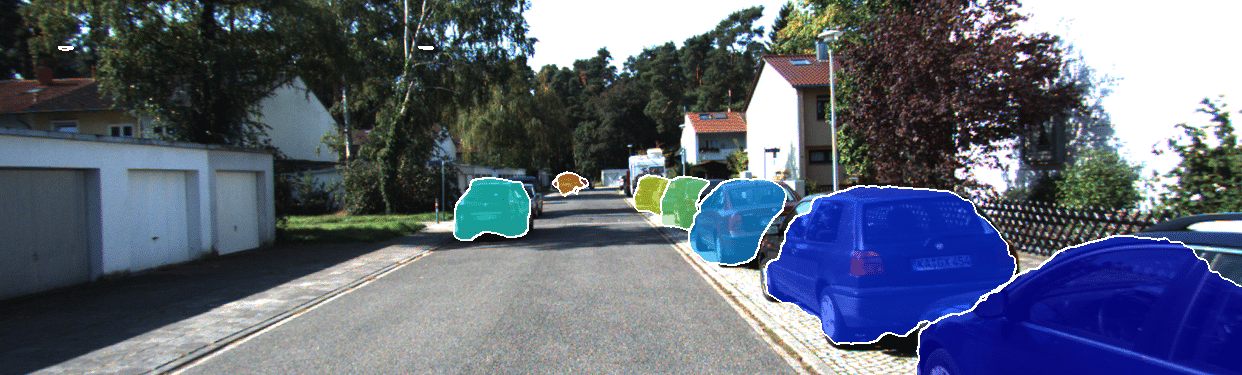}
    \includegraphics[width=\textwidth]{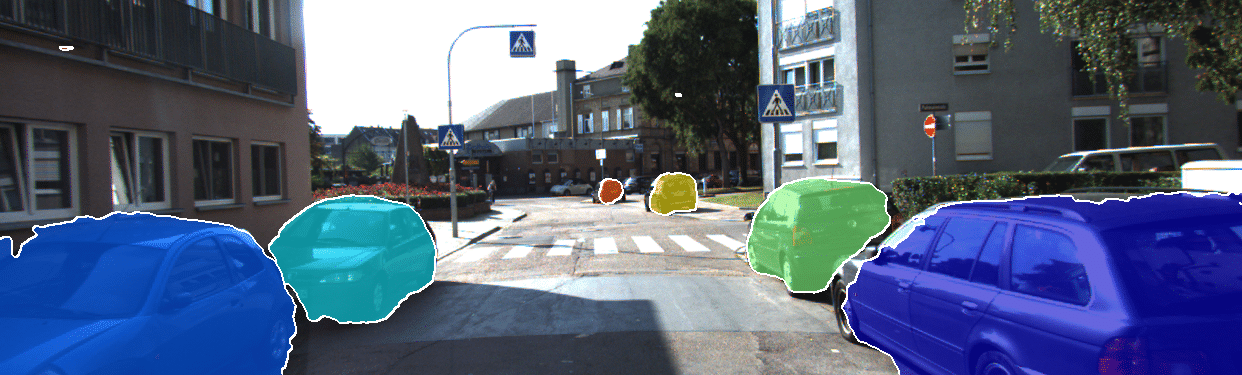}
    \includegraphics[width=\textwidth]{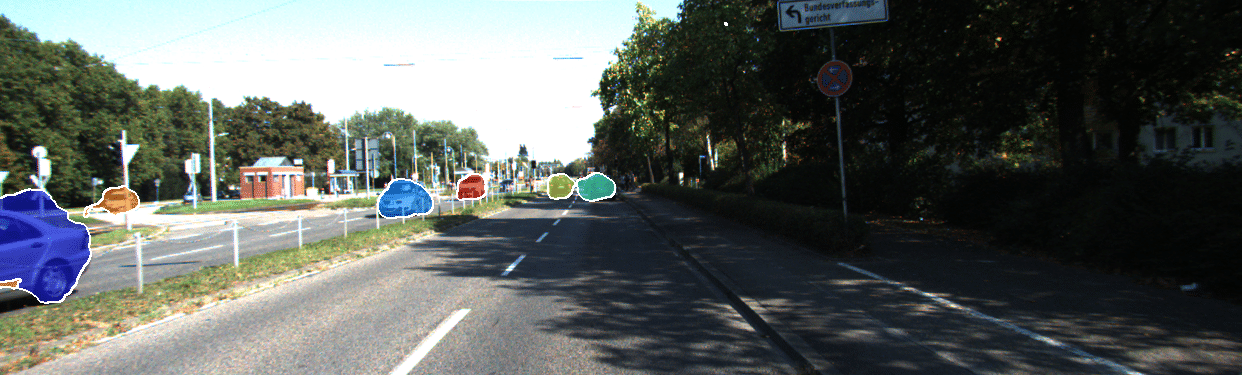}
  \end{minipage}
  \begin{minipage}{.49\textwidth}
    \includegraphics[width=\textwidth]{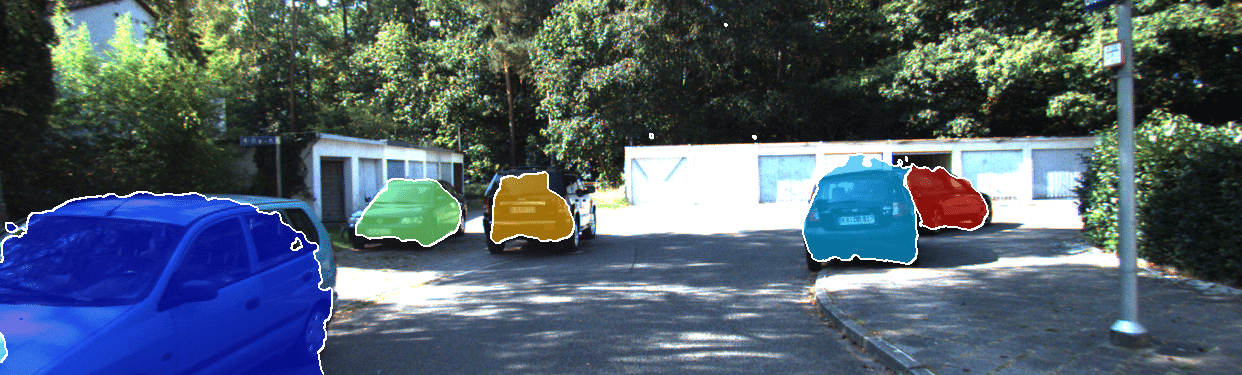}
    \includegraphics[width=\textwidth]{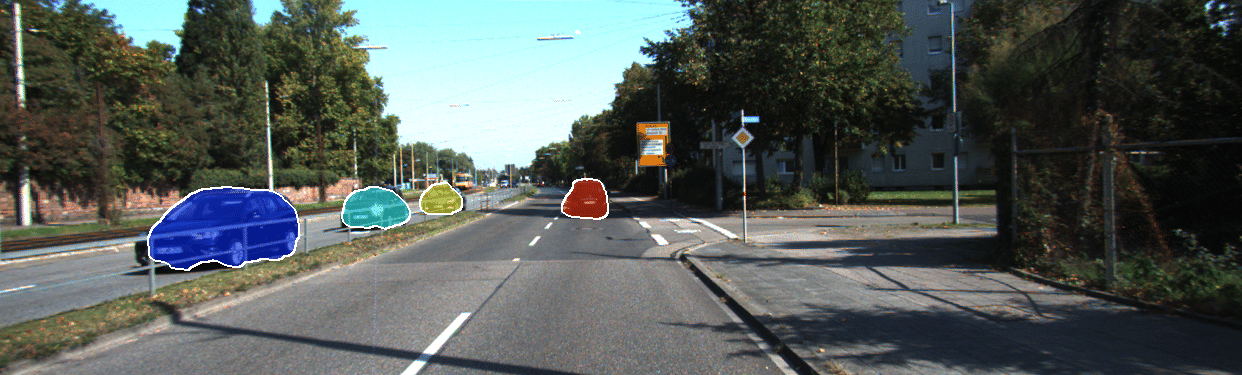}
    \includegraphics[width=\textwidth]{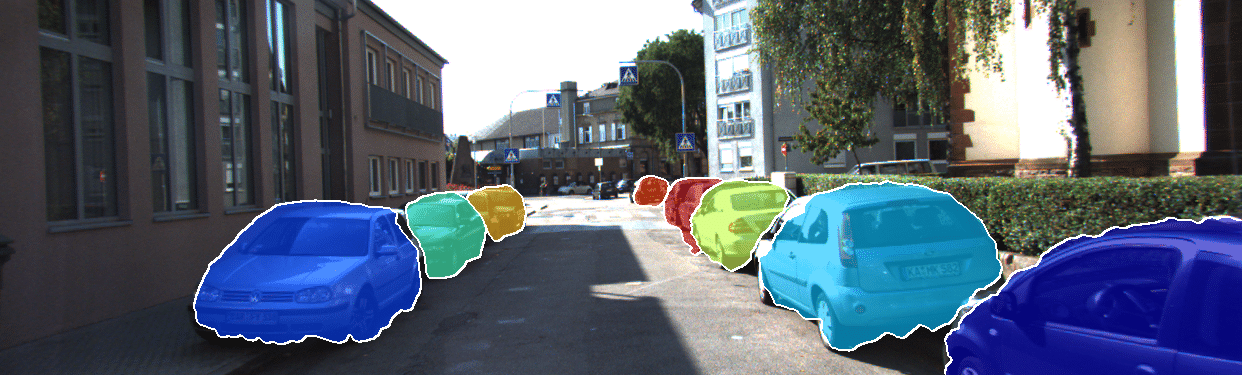}
    \includegraphics[width=\textwidth]{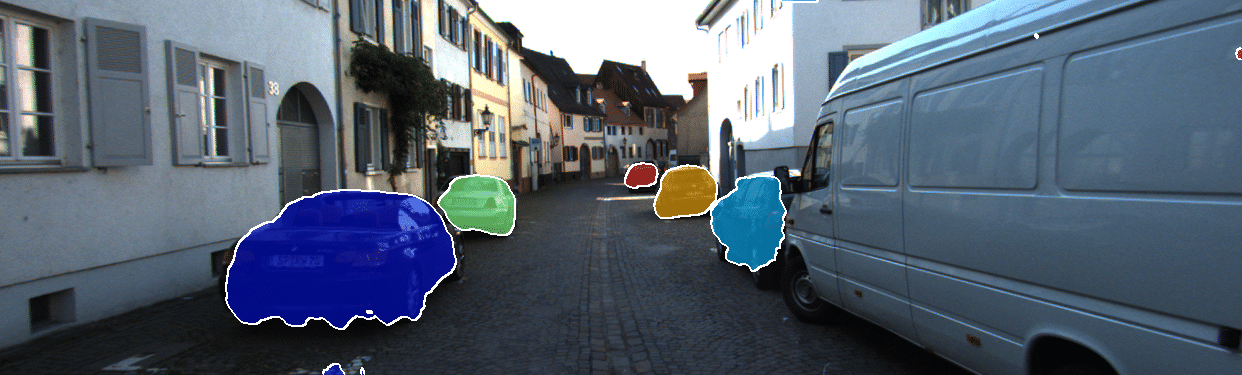}
  \end{minipage}
  \caption{Predictions of our model on the KITTI validation set}
  \label{figure:results_kitti_val}
\end{subfigure}
\caption{\emph{Predictions from our AC model on the CVPPP~(\protect\subref{figure:results_cvppp}) and KITTI datasets~(\protect\subref{figure:results_kitti_val}).} The colourmap (middle) encodes the prediction order and ranges from blue (first prediction) to red (last prediction). Note how the prediction order follows a consistent pattern: large unoccluded segments tend to be segmented first, whereas small and occluded segments are usually predicted last.}
\end{figure*}

We compare our method to other approaches on two standard instance segmentation benchmarks, each containing a rich variety of small segments as well as occlusions.

\myparagraph{CVPPP dataset.} 
For the CVPPP dataset used in our ablation study, this time we evaluate on the official 33 test images and train only our actor-critic model (\emph{AC-Dice}) on the total 128 images in the training set.

The results on the test set in \cref{table:results_cvppp} show that our method is on par with the state of the art in terms of counting while maintaining competitive segmentation accuracy.
From a qualitative analysis, see examples in \cref{figure:results_cvppp}, we observe that the order of prediction follows a consistent, interpretable pattern: large leaves are segmented first, whereas small and occluded leaves are segmented later.
This follows our intuition for an optimal processing sequence: ``easy'', more salient instances should be predicted first to alleviate consecutive predictions of the ``harder'' ones.
We also note, however, that the masks miss some fine details, such as the stalk of the leaves, which limits the benefits of the context for occluded instances.
We believe this stems from the limited capacity of the critic network to approximate a rather complex reward function.

\myparagraph{KITTI benchmark.} We use the instance-level annotation of cars in the KITTI dataset~\cite{Geiger2012CVPR} to test the scalability of our method to traffic scenes.
We used the same data split as in previous work~\cite{ren2016end,uhrig2016pixel}, which provides 3712 images for training, 144 images for validation, and 120 images for testing.
While the validation and test sets have high-quality annotations \cite{chen2014beat,ZhangFU16}, the ground-truth masks in the training set are largely ($> 95\%$) coarse or incomplete~\cite{chen2014beat}.
Hence, good generalisation from the training data would indicate that the algorithm can cope well with inaccurate ground-truth annotation.

\begin{table}[t]
\centering
\footnotesize
\subcaptionbox{KITTI test set\label{table:results_kitti_test}}{
\begin{tabularx}{\columnwidth}{@{}X@{\hspace{1em}}c@{\hspace{0.8em}}c@{\hspace{0.8em}}c@{\hspace{0.8em}}c@{}}
\toprule
Model & MWCov $\uparrow$ & MUCov $\uparrow$ & AvgFP $\downarrow$ & AvgFN $\downarrow$ \\
\midrule
DepthOrder~\cite{mono2015depth} & 70.9 & 52.2 & 0.597 & 0.736\\
DenseCRF~\cite{ZhangFU16} & 74.1 & 55.2 & 0.417 & 0.833\\
AngleFCN+D~\cite{uhrig2016pixel} & 79.7 & 75.8 & 0.201 & 0.159\\
E2E~\cite{ren2016end} & 80.0 & 66.9 & 0.764 & 0.201\\
\midrule
Ours (BL-Trunc) & 72.2 & 50.7 & 0.393 & 0.432\\
Ours (AC-IoU) & 75.6 & 57.3  & 0.338 & 0.309\\
\bottomrule
\end{tabularx}
}\\
\medskip
\subcaptionbox{KITTI validation set\label{table:results_kitti_val}}{
\begin{tabularx}{\columnwidth}{@{}X@{\hspace{1em}}c@{\hspace{1em}}c@{\hspace{1em}}c@{\hspace{1em}}c@{}}
\toprule
Model & MWCov $\uparrow$ & MUCov $\uparrow$ & AvgFP $\downarrow$ & AvgFN $\downarrow$ \\
\midrule
E2E (Iter-1) & 64.1 & 54.8 & 0.200 & 0.375\\
E2E (Iter-3) & 71.3 & 63.4 & 0.417 & 0.308\\
E2E (Iter-5) & 75.1 & 64.6 & 0.375 & 0.283\\
\midrule
Ours (BL-Trunc) & 70.4 & 55.8 & 0.313 & 0.339\\
Ours (AC-IoU) & 71.9 & 59.5  & 0.262 & 0.253\\
\bottomrule
\end{tabularx}
}
\caption{\emph{Segmentation quality on KITTI.} We evaluate our baseline with truncated BPTT (\textit{BL-Trunc}) and the actor-critic with IoU-based reward (\textit{AC-IoU}) in terms of mean weighted (MWCov) and unweighted (MUWCov) coverage, average false positive (AvgFP), and false negative (AvgFN) rates.}
\vspace{-0.5em}
\end{table}

The evaluation criteria for this dataset are: the mean weighted coverage loss (MWCov), the mean unweighted coverage loss (MUCov), the average false positive rate (AvgFP), and the average false negative rate (AvgFN).
MUCov is the maximum IoU of the ground truth with a predicted mask, averaged over all ground-truth segments in the image.
MWCov additionally weighs the IoUs by the area of the ground-truth mask.
AvgFP is the fraction of mask predictions that do not overlap with the ground-truth segments.
Conversely, AvgFN measures the fraction of the ground-truth segments that do not overlap with the predictions.

We use an IoU-based score function to compute the rewards, \ie $\mathcal{F}(\mathcal{S}, \mathcal{T}) = \frac{\sum_i{\mathcal{S}_i \mathcal{T}_i}}{\sum_i \mathcal{S}_i + \sum_i \mathcal{T}_i - \sum_i{\mathcal{S}_i \mathcal{T}_i}}$.
To show the benefits of our Actor-Critic model (\emph{AC-IoU}) for structured prediction at higher resolutions, we also train and report results for a baseline, the actor-only model trained with one step BPTT (\emph{BL-Trunc}).
Considering the increased variability of the dataset compared to CVPPP, we used 64 latent dimensions for the action space.

The results on the test are shown in Table~\ref{table:results_kitti_test}.
Given the relatively small size of the test set, we also report the results on the validation set in Table~\ref{table:results_kitti_val}, and use the available results from the equivalent evaluation of a state-of-the-art method~\cite{ren2016end} for reference.

The results indicate that our method scales well to larger resolutions and action spaces and shows competitive accuracy despite not using bounding box representations.
Similar to our results on CVPPP, our model does not quite reach the accuracy of a recurrent model using bounding boxes~\cite{ren2016end} and a non-recurrent pipeline.
We believe the segmentation accuracy is currently limited by the degree of the reward approximation by the critic and the representational power of the network architecture used by the actor model.
As can be seen in some examples in \cref{figure:results_kitti_val}, without post-processing the masks are not always well aligned with the object and occlusion boundaries.
However, we note that the prediction order also follows a consistent, interpretable pattern: nearby instances are segmented first, while far-away instances are segmented last.
Without hard-coding such constraints, the network appears to have learned a strategy that agrees with human intuition to segment larger, close-by objects first and exploits the resulting context to make predictions in the order of increasing difficulty. 

\section{Conclusions}

In the current study, we formalised the task of instance segmentation in the framework of reinforcement learning.
Our proposed actor-critic model utilises exploration noise to alleviate the initialisation bias on the prediction ordering.
Considering the high dimensionality of pixel-level actions, we enabled exploration in the action space by learning a low-dimensional representation through a conditional variational auto-encoder.
Furthermore, the critic approximates a reward signal that also accounts for future predictions at any given timestep.  
In our experiments, it attained competitive results on established instance segmentation benchmarks and showed improved segmentation performance at the later timesteps.
Our model predicts instance masks directly at the full resolution of the input image and does not require intermediate bounding box predictions,
which stands in contrast to proposal-based architectures~\cite{he2017mask} or models delivering only a preliminary representation for further post-processing, \eg~ \cite{de2017semantic,uhrig2016pixel}.

These encouraging results suggest that actor-critic models have potentially a wider application spectrum,
since the critic network was able to learn a rather complex loss function to a fair degree of approximation.
In future work, we aim to improve our baseline model of the actor network, which currently limits the attainable accuracy.

{\small
\myparagraph{Acknowledgements.} The authors would like to thank Stephan R.~Richter for helpful discussions.
}

{
\small
\balance
\bibliographystyle{ieee_fullname}
\bibliography{egbib}
}

\clearpage

\pagenumbering{roman}
\appendix

{\twocolumn[\begin{@twocolumnfalse}
{\newpage
\null
\vskip .375in
\begin{center}
   {\Large \bf \thesupptitle \par}
   \vspace*{24pt}
   {
   \large
   \lineskip .5em
   \begin{tabular}[t]{c}
      \thesuppauthor
   \end{tabular}
   \par
   }
   \vskip .5em
   \vspace*{12pt}
\end{center}
}\end{@twocolumnfalse}]}

\nobalance

\section{Ordering of Prediction}
As we discussed in the main text, previous work suggests that the overall accuracy of a recurrent model is not invariant to the prediction order~\citesupp{li2017not,OrderMatters}.
To confirm this experimentally for instance segmentation, we decouple the localisation and the segmentation aspect of a recurrent model in an oracle experiment.
Reserving the location of the ground-truth segments as the oracle knowledge (\eg, with a bounding box) allows us to control the prediction order and study its role for the quality of the pixelwise prediction.

\begin{figure}[t]
\centering
\includegraphics[width=0.33\textwidth]{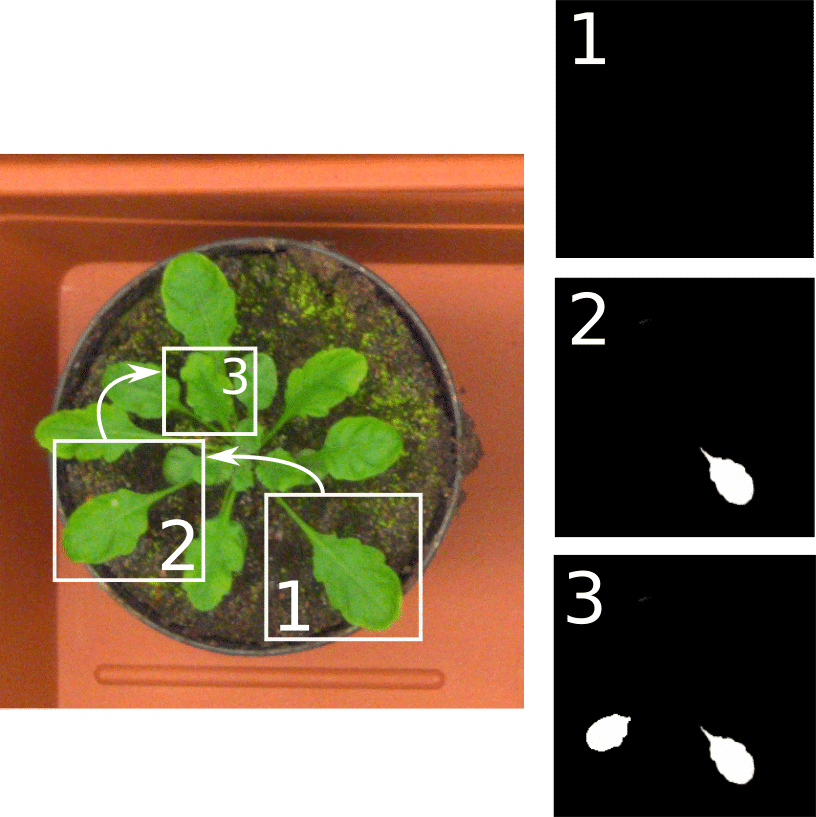}
\caption{\emph{Oracle experiment:} We extract patches of the image and the context mask centred on the ground-truth instances and supply only those to the segmentation network.
The oracle knowledge about the object location allows us to focus exclusively on the pixelwise prediction within the bounding box, given a specific context of occlusions and the previously predicted segments.}
\label{figure:oracle_permutation}
\end{figure}

The setup for our oracle experiment is illustrated in \cref{figure:oracle_permutation}.
We supply the image and context patches (4 input channels) to a segmentation network in random order, while at the same time accumulating its predictions into the (global) context mask.
We repeat this procedure 20 times for every image, with a different random order each.
We use the CVPPP train/val split from our ablation study, and report the results on the 25 images in the validation set.
The architecture of the network is the same as the segmentation module used by Ren and Zemel~\citesupp{ren2016end}.

\begin{figure*}[t]
\begin{subfigure}{.7\textwidth}
  \centering
  \includegraphics[width=\textwidth]{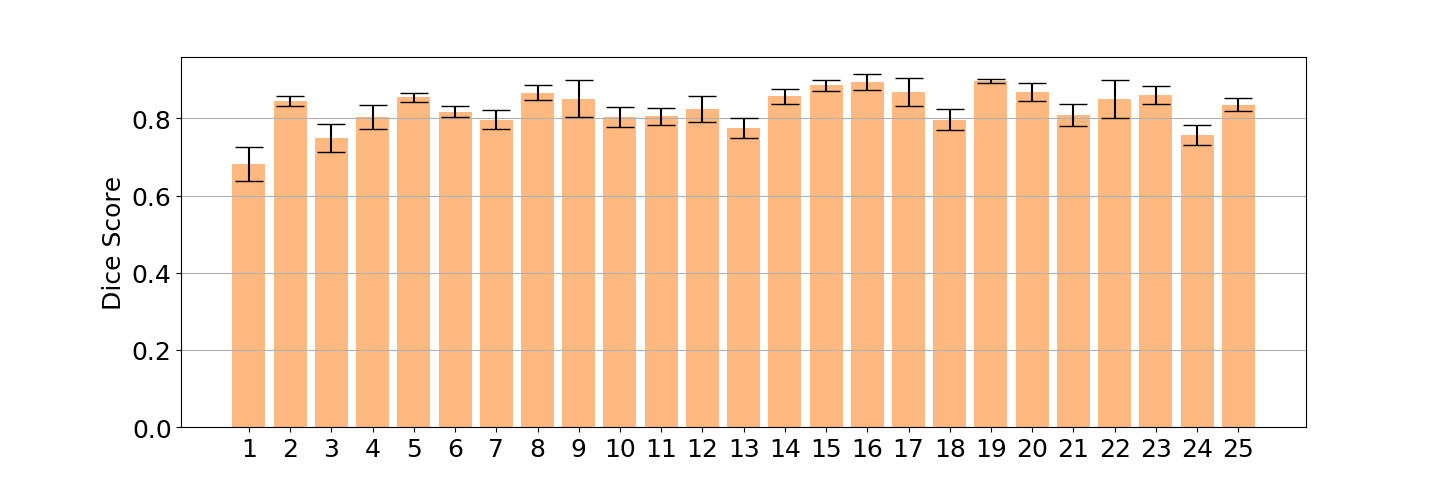}
  \caption{Mean Dice for 25 examples from CVPPP (val)}
  \label{figure:oracle_hist}
\end{subfigure}
\begin{minipage}{.3\textwidth}
\begin{subfigure}{\textwidth}
  \centering
  \includegraphics[width=0.375\textwidth]{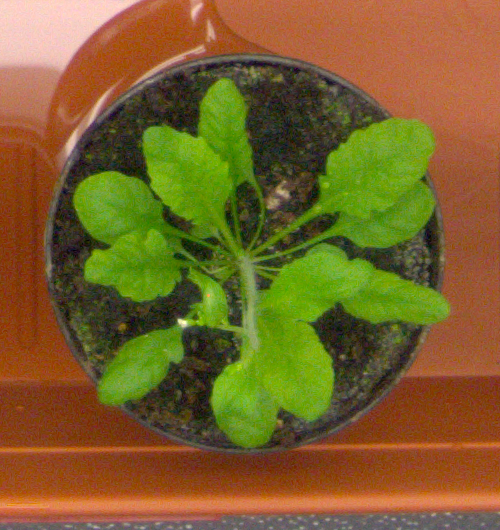}  \hspace{0.5em}  \includegraphics[width=0.4\textwidth]{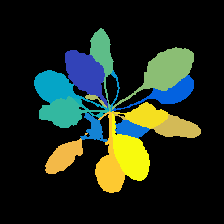}
  \caption{Example $\#1$}
  \label{figure:oracle_std_largest}
\end{subfigure}\\[1mm]
\begin{subfigure}{\textwidth}
  \centering
  \includegraphics[width=0.375\textwidth]{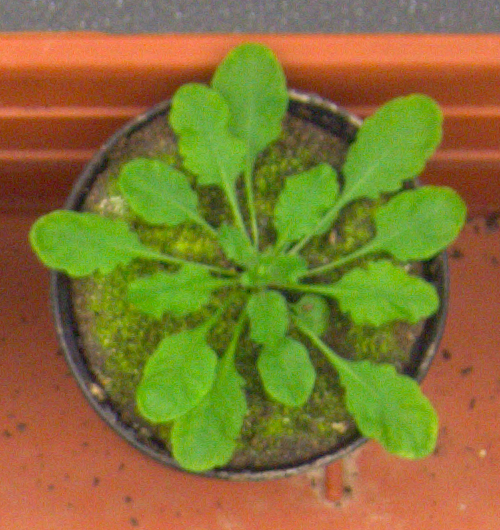}  \hspace{0.5em}  \includegraphics[width=0.4\textwidth]{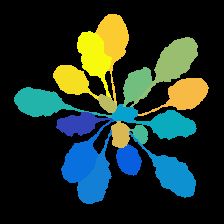}
  \caption{Example $\#19$}
  \label{figure:oracle_std_smallest}
\end{subfigure}
\end{minipage}
\caption{\emph{Prediction of ground-truth segments in different order by using oracle bounding boxes centred on each of the target segments:}
\emph{(\protect\subref{figure:oracle_hist})} Mean Dice and standard deviation (error bars) of sequential mask prediction from 20 random permutations of oracle bounding boxes.
The best and the worst prediction sequence averaged across the images yield $86.5$ and $78.1$ Dice, which implies the potential benefit of the optimal prediction ordering.
\emph{(\protect\subref{figure:oracle_std_largest})} Example from CVPPP with the highest standard deviation, \ie which can particularly benefit from the optimal ordering.
\emph{(\protect\subref{figure:oracle_std_smallest})} Example from CVPPP with the lowest standard deviation, \ie where ordering of prediction has little effect on Dice.
Note how the benefits of choosing an optimal ordering particularly occur in scenes with occlusions.
}
\label{figure:oracle_ordering} 
\end{figure*}

Figure~\ref{figure:oracle_hist} shows the statistic of the results (mean and standard deviation) for each of the 25 images.
We observe that there is a tangible difference between the different segmentation orderings.
The mean of the maximum and the minimum Dice across the images are $86.5$ and $78.1$ Dice, respectively, which corresponds to a potential gain of $8.4$ mean Dice for the optimal prediction order.
Notably, the deviation between the runs varies depending on the image.
To investigate this further, we show two examples from the validation set with the highest (index \#1, \cref{figure:oracle_std_largest}) and the lowest (index \#19, \cref{figure:oracle_std_smallest}) deviation.
Example \#1 appears to contain many more occlusions and more complex segment shapes (\eg, a single stalk) compared to example \#19.
This aligns well with our expectation: the benefit of the context should be more pronounced in more complex scenes.

\section{Architecture}
The architecture of the actor network and its parameters depending on the dataset used are shown in \cref{table:actor_arch,table:actor_params}.
The critic has a similar architecture, but makes use of batch normalisation (BN)~\citesupp{IoffeS15}.
Additionally, one fully-connected layer (FC) from the last convolution is replaced by global maximum pooling.
Further details are summarised in Table~\ref{table:critic_arch}. 

We use a vanilla LSTM~\citesupp{graves2005framewise} without peephole connections to represent the hidden state due to its established performance~\citesupp{greff2017lstm} and more economic use of memory compared to convolutional 
LSTMs~\citesupp{romera2016recurrent,xingjian2015convolutional}.

The pre-processing network, which predicts the foreground and the angle quantisation of instances, is an FCN~\citesupp{long2015fully} with the same architecture as used by Ren and Zemel~\citesupp{ren2016end}.

\section{Training}
To train our actor-critic model, we empirically found that gradually increasing the maximum sequence length leads to faster convergence than training on the complete length from the start.
Specifically, we train the actor to predict $n$ remaining masks, where $n$ starts with $1$ and is gradually extended until the maximum number of instances in the dataset plus the terminations state (21 for CVPPP and 16 for KITTI) is reached.
For example, with $n=1$ the model learns to predict the last instance, \ie the initial state $s_1$ input to the network is an accumulation of all but one ground-truth masks on a given image, which we choose randomly.
We extend $n$ by 5 once the segmentation accuracy on the validation set (either Dice or IoU) stops improving for several epochs.

We decrease the learning rate by a factor of 10 if we observe either no improvement or high variance of the results on the validation set.
To trade-off the critic's gradient, we used a constant scalar $\beta_\text{act}=10^{-3}$ for the KL-divergence loss.
The actor and critic networks use weight decay regularisation of $10^{-5}$ and $10^{-4}$, respectively.
We use Adam~\citesupp{kingma14adam} for optimisation, as our experiments with SGD lead to considerably slower convergence.
To manage training time, we downscale the original images with bilinear interpolation.
The ground-truth masks are scaled down using nearest-neighbour interpolation.
For CVPPP, we reduce the original resolution from $530 \times 500$ to $224 \times 224$, and for KITTI from the original $1242 \times 375$ to $768 \times 256$.
We implemented our method in Lua Torch~\citesupp{Collobert2011Torch7AM}.
Both pre-training and training were performed on a single Nvidia Titan X GPU.
The code is available at \mbox{\url{https://github.com/visinf/acis/}}.

\section{Qualitative Examples}
\begin{figure*}[t]
\centering
  \begin{minipage}{.15\textwidth}
  \centering
  	$\text{Input image}$
    \includegraphics[width=0.91\textwidth]{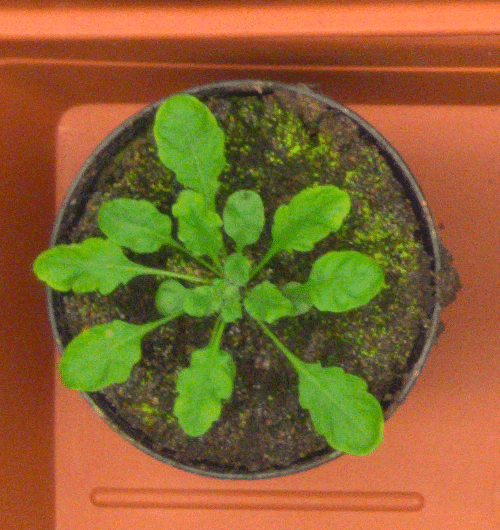}
  \end{minipage}
  \begin{minipage}{.15\textwidth}
  	$\text{t = 1}$
    \includegraphics[width=\textwidth]{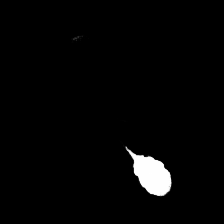}
  \end{minipage}
  \begin{minipage}{.15\textwidth}
    $\text{t = 2}$
    \includegraphics[width=\textwidth]{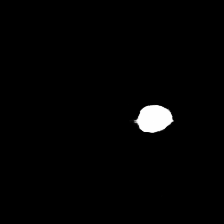}
  \end{minipage}
  \begin{minipage}{.15\textwidth}
  $\text{t = 3}$
    \includegraphics[width=\textwidth]{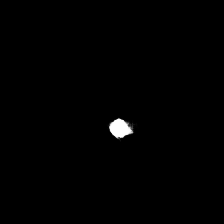}
  \end{minipage}
    \begin{minipage}{.15\textwidth}
    $\text{t = 4}$
    \includegraphics[width=\textwidth]{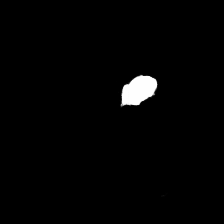}
  \end{minipage}
    \begin{minipage}{.15\textwidth}
    $\text{t = 5}$
    \includegraphics[width=\textwidth]{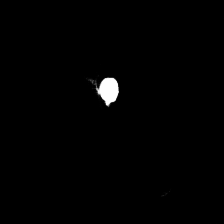}
  \end{minipage}\\
  \vspace{1em}
  \begin{minipage}{.15\textwidth}
      $\text{t = 6}$
    \includegraphics[width=\textwidth]{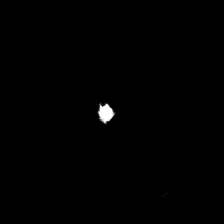}
  \end{minipage}
      \begin{minipage}{.15\textwidth}
      $\text{t = 7}$
    \includegraphics[width=\textwidth]{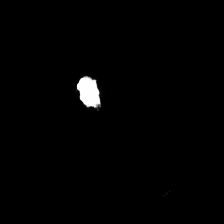}
  \end{minipage}
      \begin{minipage}{.15\textwidth}
      $\text{t = 8}$
    \includegraphics[width=\textwidth]{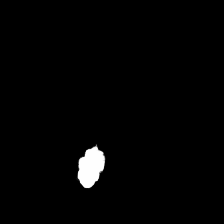}
  \end{minipage}
      \begin{minipage}{.15\textwidth}
      $\text{t = 9}$
    \includegraphics[width=\textwidth]{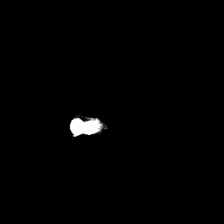}
  \end{minipage}
      \begin{minipage}{.15\textwidth}
      $\text{t = 10}$
    \includegraphics[width=\textwidth]{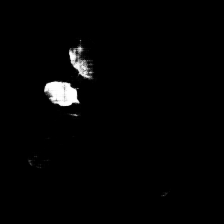}
  \end{minipage}
      \begin{minipage}{.15\textwidth}
      $\text{t = 11}$
    \includegraphics[width=\textwidth]{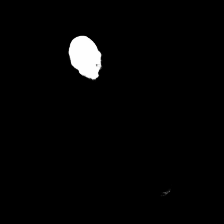}
  \end{minipage}\\
  \vspace{1em}
   \begin{minipage}{.15\textwidth}
      $\text{t = 12}$
    \includegraphics[width=\textwidth]{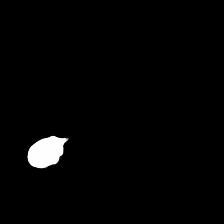}
  \end{minipage}
      \begin{minipage}{.15\textwidth}
      $\text{t = 13}$
    \includegraphics[width=\textwidth]{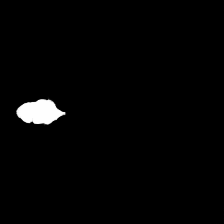}
  \end{minipage}
      \begin{minipage}{.15\textwidth}
      $\text{t = 14}$
    \includegraphics[width=\textwidth]{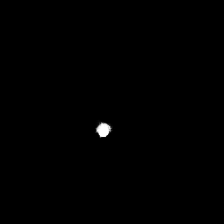}
  \end{minipage}
      \begin{minipage}{.15\textwidth}
      $\text{t = 15}$
    \includegraphics[width=\textwidth]{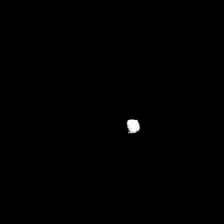}
  \end{minipage}
        \begin{minipage}{.15\textwidth}
        $\text{t = 16}$
    \includegraphics[width=\textwidth]{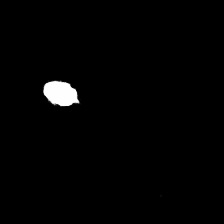}
  \end{minipage}
        \begin{minipage}{.15\textwidth}
        $\text{t = 17}$
    \includegraphics[width=\textwidth]{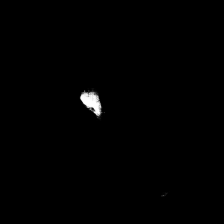}
  \end{minipage}\\
  \vspace{1em}
  \begin{minipage}{.15\textwidth}
  	\centering
  	$\text{Result}$
    \includegraphics[width=\textwidth]{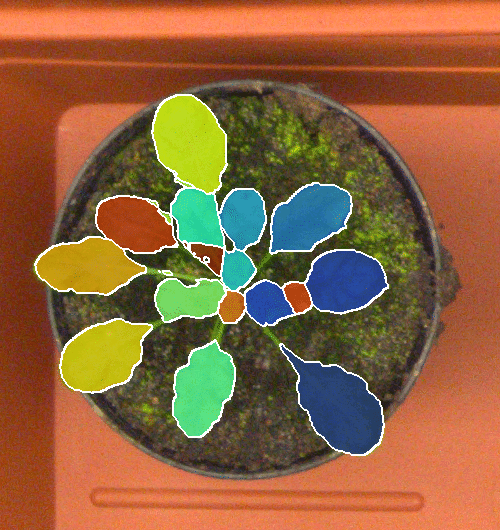}
  \end{minipage}
  \caption{\emph{Visualisation of individual masks as they are predicted at each timestep $t$ on CVPPP val.} Interestingly, our model continued to predict quality masks despite an inaccurate prediction at timestep $t=10$.}
  \label{figure:cvppp_detailed}
\end{figure*}

\begin{figure*}[t]
\centering
  \begin{minipage}{.33\textwidth}
  \centering
  	$\text{Input image}$
    \includegraphics[width=\textwidth]{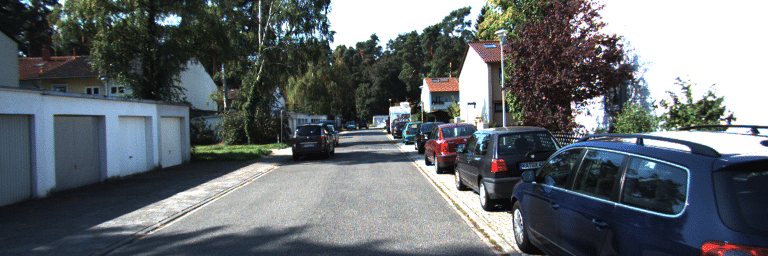}
  \end{minipage}
  \begin{minipage}{.33\textwidth}
  	$t = 1$\\
    \includegraphics[width=\textwidth]{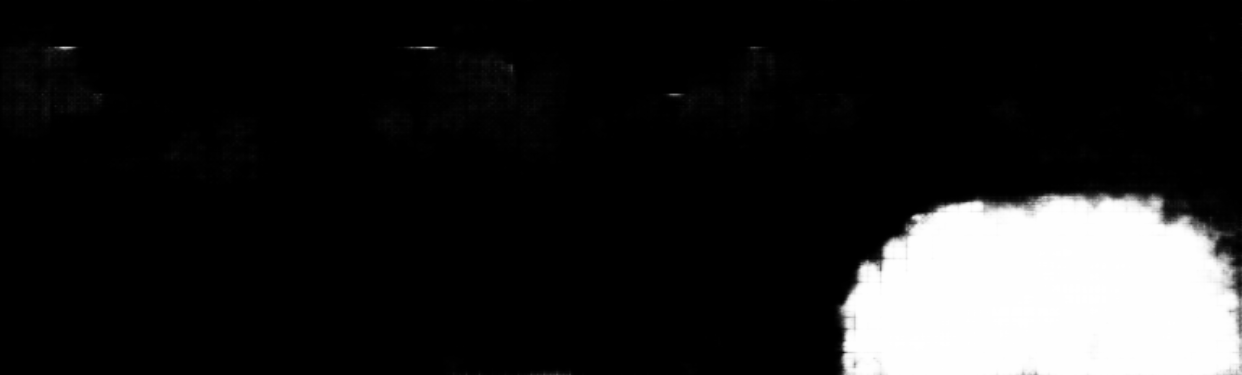}
  \end{minipage}
  \begin{minipage}{.33\textwidth}
    $\text{t = 2}$\\
    \includegraphics[width=\textwidth]{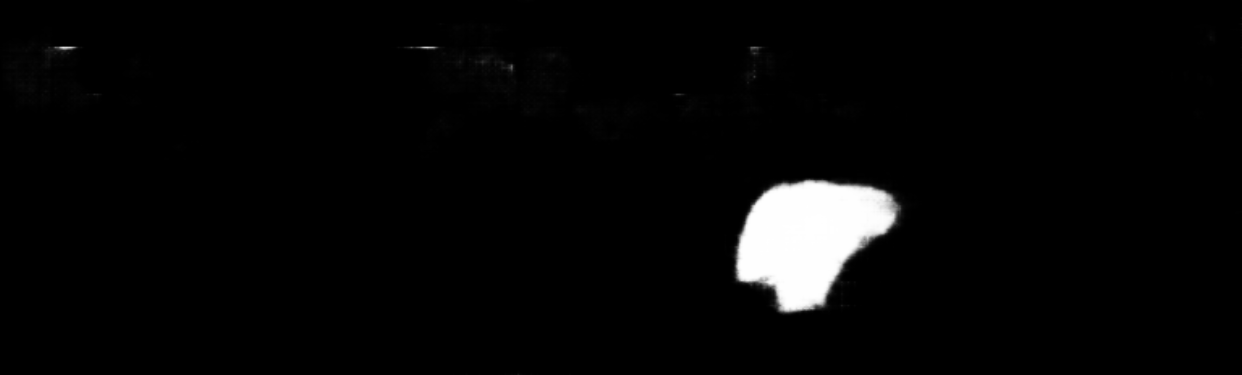}
  \end{minipage}\\
  \vspace{1em}
  \begin{minipage}{.33\textwidth}
  $\text{t = 3}$\\
    \includegraphics[width=\textwidth]{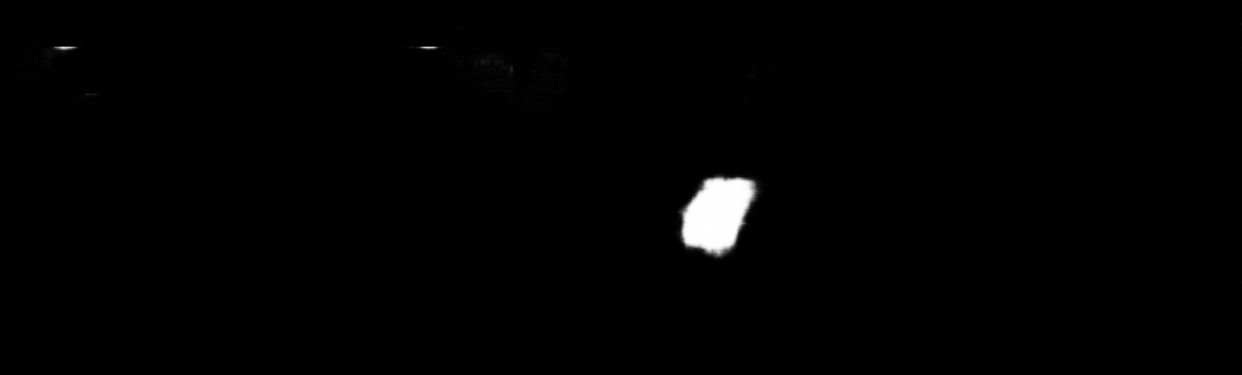}
  \end{minipage}
    \begin{minipage}{.33\textwidth}
    $\text{t = 4}$\\
    \includegraphics[width=\textwidth]{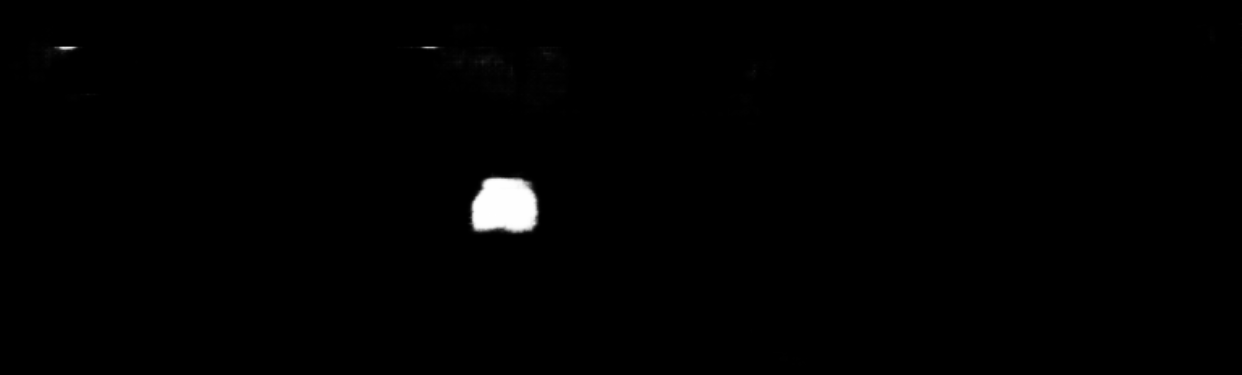}
  \end{minipage}
    \begin{minipage}{.33\textwidth}
    $\text{t = 5}$\\
    \includegraphics[width=\textwidth]{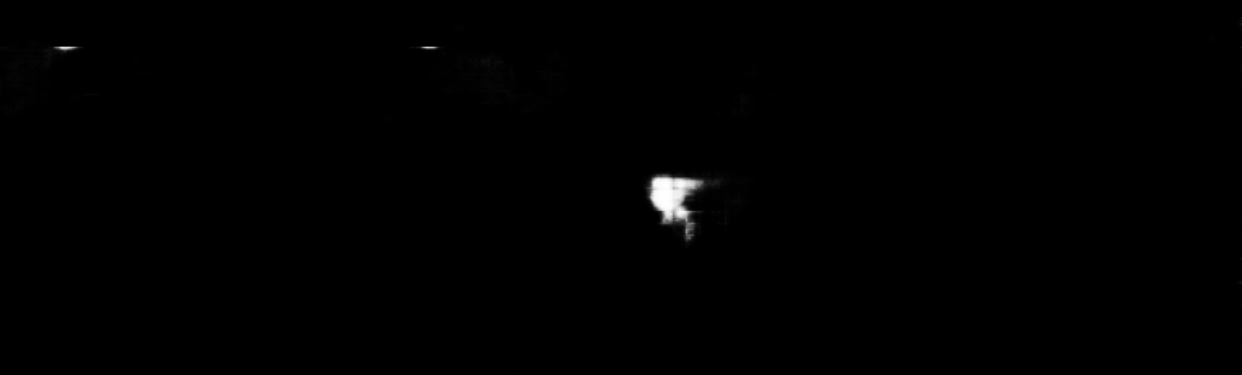}
  \end{minipage}\\
  \vspace{1em}
  \begin{minipage}{.33\textwidth}
      $\text{t = 6}$\\
    \includegraphics[width=\textwidth]{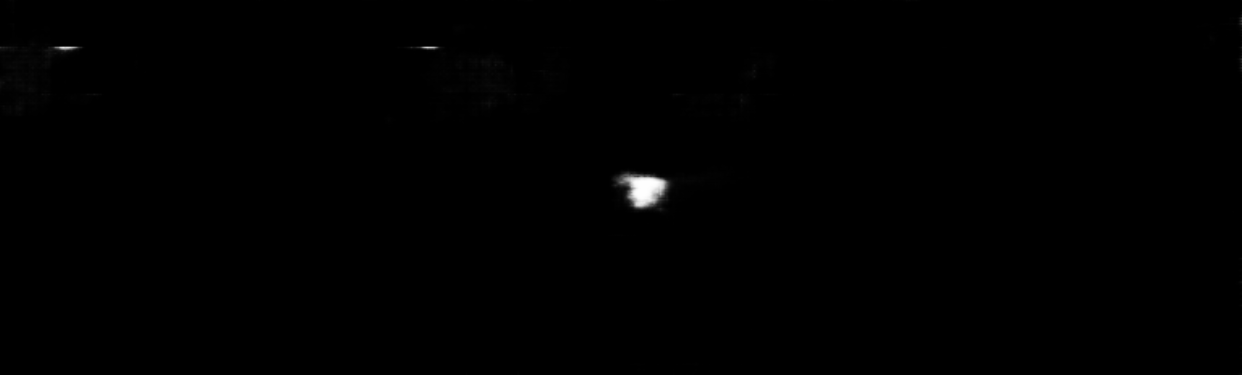}
  \end{minipage}
      \begin{minipage}{.33\textwidth}
      $\text{t = 7}$\\
    \includegraphics[width=\textwidth]{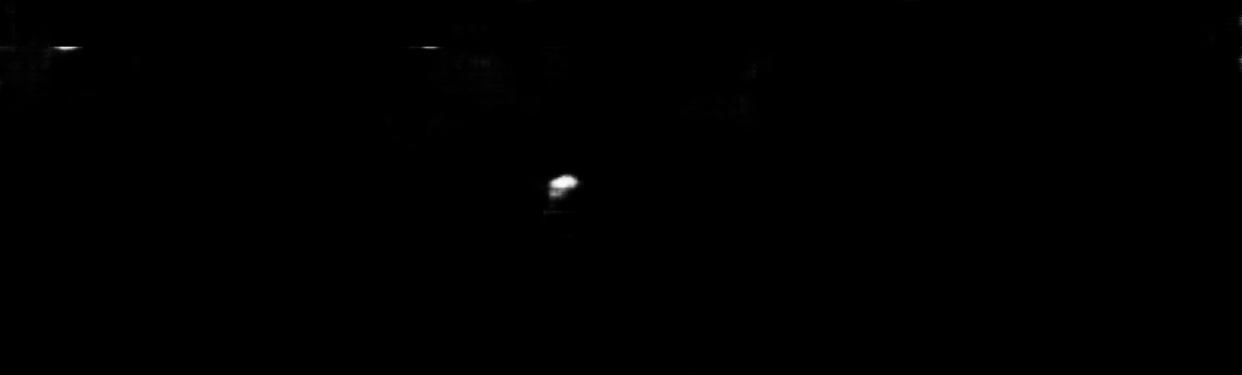}
  \end{minipage}
      \begin{minipage}{.33\textwidth}
      $\text{t = 8}$\\
    \includegraphics[width=\textwidth]{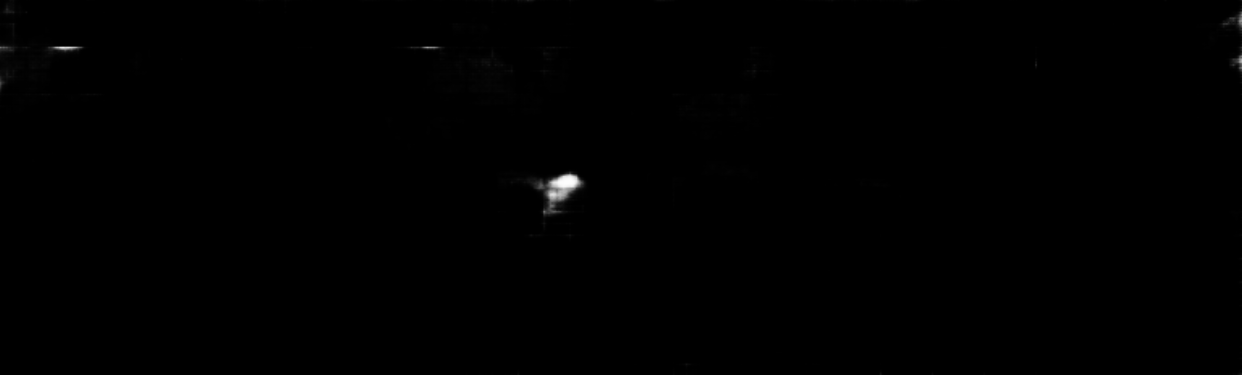}
  \end{minipage}\\
  \vspace{1em}
  \begin{minipage}{.33\textwidth}
  	\centering
  	$\text{Result}$
    \includegraphics[width=\textwidth]{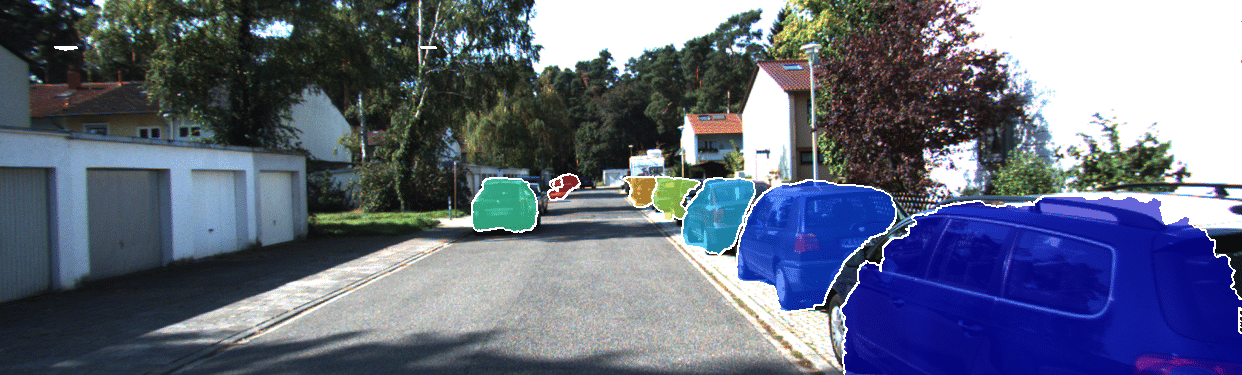}
  \end{minipage}
  \caption{\emph{Visualisation of individual masks as they are predicted at each timestep $t$ on KITTI val.} We observe that the prediction order in this case strongly correlates with the vicinity of the vehicles to the camera.}
  \label{figure:kitti_detailed}
\end{figure*}

We provide qualitative segment-by-segment visualisations in \cref{figure:cvppp_detailed,figure:kitti_detailed} from the CVPPP and KITTI validation sets.
Supporting our analysis in the main paper, the order of prediction exhibits a consistent pattern.
Furthermore, we observe that our model copes well with inaccuracies of intermediate predictions -- a common failure mode of recurrent networks~\citesupp{Ranzato:2016:SLT}.

\begin{table*}[!t]
\centering
\begin{tabular}{@{}l|cccc@{}}
\toprule
Section & Type & (Kernel) size & Stride & \# of channels\\
\midrule
\multirow{10}{*}{Encoder} & Conv $\rightarrow$ ReLU & $3\times 3$ & 1 & 32\\
 & MaxPooling  & $2\times 2$ & 2 & --\\
 & Conv $\rightarrow$ ReLU & $3\times 3$ & 1 & 48\\
 & MaxPooling  & $2\times 2$ & 2 & --\\
 & Conv $\rightarrow$ ReLU & $3\times 3$ & 1 & 64\\
 & MaxPooling  & $2\times 2$ & 2 & -- \\
 & Conv $\rightarrow$ ReLU & $3\times 3$ & 1 & 96\\
 & MaxPooling  & $2\times 2$ & 2 & -- \\
 & Conv $\rightarrow$ ReLU & $3\times 3$ & 1 & 128\\
 & MaxPooling & $2\times 2$ & 2 & -- \\
\midrule
\multirow{6}{*}{Bottleneck} & FC $\rightarrow$ LeakyReLU & $128 \times h \times w$ & -- & $\text{size}(h_t)$\\
& LSTM & $\text{size}(h_t)$ & -- & $\text{size}(h_t)$ \\
& FC  $\rightarrow$ LeakyReLU & $\text{size}(h_t)$ & -- & $z$\\
& FC ($\mu$, $\sigma$) & $z$ & -- & $2 \times l$\\
& FC $\rightarrow$ LeakyReLU & $l$ & -- & $z$\\
& FC $\rightarrow$ LeakyReLU & $z$ & -- & $128 \times h \times w$\\
\midrule
\multirow{6}{*}{Decoder} & Deconv $\rightarrow$ ReLU & $3\times 3$ & 2 & 128 + $\text{SP}_5$\\
 & Deconv $\rightarrow$ ReLU & $3\times 3$ & 2 & 96 + $\text{SP}_4$\\
 & Deconv $\rightarrow$ ReLU & $3\times 3$ & 2 & 64 + $\text{SP}_3$\\
 & Deconv $\rightarrow$ ReLU & $3\times 3$ & 2 & 48 + $\text{SP}_2$\\
 & Deconv $\rightarrow$ ReLU & $3\times 3$ & 2 & 32 + $\text{SP}_1$\\
 & Deconv $\rightarrow$ ReLU & $3\times 3$ & 1 &  1 + $\text{SP}_0$\\
\bottomrule
\end{tabular}
\caption{\emph{Architecture of the actor network.} $\text{SP}_m$ denotes the additional channels in the State Pyramid provided at resolution scaled by $2^m$.}
\label{table:actor_arch}
\end{table*}

\begin{table*}[!t]
\centering
\begin{tabular}{@{}l|cccc@{}}
\toprule
Dataset & $h \times w$ & $size(h_t)$ & $l$ & $z$\\
\midrule
CVPPP & $7 \times 7$ & $512$ & $16$ & $256$\\
KITTI & $8 \times 24$ & $512$ & $64$ & $512$\\
\bottomrule
\end{tabular}
\caption{\emph{Parameter values for the actor-critic.}}
\label{table:actor_params}
\end{table*}

\begin{table*}[!t]
\centering
\begin{tabular}{@{}l|ccc@{}}
\toprule
Type & (Kernel) size & Stride & \# of output channels\\
\midrule
 Conv $\rightarrow$ BN $\rightarrow$ ReLU & $3\times 3$ & 1 & 32\\
 MaxPooling  & $2\times 2$ & 2 & -- \\
 Conv $\rightarrow$ BN $\rightarrow$ ReLU & $3\times 3$ & 1 & 64\\
 MaxPooling  & $2\times 2$ & 2 & -- \\
 Conv $\rightarrow$ BN $\rightarrow$ ReLU & $3\times 3$ & 1 & 128\\
 MaxPooling  & $2\times 2$ & 2 & -- \\
 Conv $\rightarrow$ BN $\rightarrow$ ReLU & $3\times 3$ & 1 & 256\\
 MaxPooling  & $2\times 2$ & 2 & -- \\
 Conv $\rightarrow$ BN $\rightarrow$ ReLU & $3\times 3$ & 1 & 512\\
 Global MaxPooling & $h \times w$ & 1 & 512 \\
 FC $\rightarrow$ LeakyReLU & 512 & -- & 1024\\
 FC $\rightarrow$ LeakyReLU & 1024 & -- & 1024\\
 FC $\rightarrow$ LeakyReLU & 1024 & -- & 512\\
 FC & 512 & -- & 1\\
\bottomrule
\end{tabular}
\caption{\emph{Architecture of the critic network.}}
\label{table:critic_arch}
\end{table*}

{
\small
\balance
\bibliographystylesupp{ieee_fullname}
\bibliographysupp{egbib_supp}
}

\end{document}